\pdfoutput=1
\documentclass[letterpaper]{article} 
\usepackage[preprint]{aaai2027}
\usepackage[hyphens]{url}  
\usepackage{graphicx} 
\usepackage{natbib}  
\usepackage{caption} 
\usepackage{booktabs}
\newenvironment{promptbox}[1]{\par\addvspace{8pt}\noindent\rule{\linewidth}{0.8pt}\par\nobreak\vspace{3pt}%
  \noindent{\bfseries #1}\par\nobreak\vspace{2pt}\noindent\rule{\linewidth}{0.4pt}\par\nobreak\vspace{5pt}%
  \begingroup\ttfamily\footnotesize\raggedright\setlength{\parindent}{0pt}\setlength{\parskip}{2.5pt}}%
  {\par\endgroup\vspace{4pt}\noindent\rule{\linewidth}{0.8pt}\par\addvspace{8pt}}
\newcommand{\promptsec}[1]{{\bfseries #1}}
\title{PAVE: A Harness for Geometrically Consistent Road Network Modeling with Vision-Language Models}
\author{
    Yajun Liu,
    Jinxi Wu,
    Juan Pablo Mill\'{a}n Raya,
    Xuesong Zhou
}
\affiliations{
    Arizona State University\\
    yajunliu@asu.edu, jinxiwu@asu.edu, jmillanr@asu.edu, xzhou74@asu.edu
}

\begin{document}
\maketitle

\begin{abstract}
Transportation models, whether they simulate traffic or support planning, are only as reliable as the road networks they run on, down to the lanes and turns at each intersection. Practitioners still take such a network from GIS files and supply its junction and movement data by hand, a task that recurs with each new base year and design alternative, or they generate it from open map data. OpenStreetMap's lane tags give the correct layout for only 13 of the 44 approaches we test; open map data are often incomplete or inaccurate. Satellite and street-level imagery, in contrast, cover all 44 and show their lanes as painted. Vision-language models (VLMs) can read such imagery, but asked directly they lose geometric consistency when the same lane must be matched across a satellite crop, street-level photos taken from different poses and an ID-labelled network. We therefore present PAVE (Perceive, Anchor, Verify, Edit), a harness in which code measures what can be measured, traffic engineering knowledge enters as general rules and checks, and the model makes only the open judgements, each anchored to network IDs so that checks catch answers that contradict a measurement, a second reading or a traffic rule. Scored against human labels at signalised intersections, PAVE raises the share of exactly correct lane layouts from 49--64\% with a single request to 86--91\% for two VLMs, with most of the gain coming from reading one approach at a time on traced lanes. The same division of work edits a real network to a planning brief: the model names the links and nodes of each part of a facility, and code builds the edit from a design template. Of 18 edits, 16 pass every network check, against none when the model writes its own code, and 2 contain a design or coding error, against 7. An analysis shows where VLMs lose geometric consistency and which check catches each failure.
\end{abstract}

\section{Introduction}
\begin{figure*}[t]
\centering
\includegraphics[width=\textwidth]{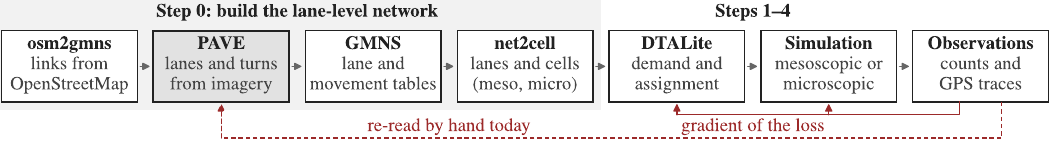}
\caption{Building the network is step 0 of a transportation model, here with open tools on GMNS. Red: the gaps return to steps 1--4 as gradients \citep{wu2018hierarchical}, and to step 0 only manually (dashed); no backward pass is run here.}
\label{fig:stepzero}
\end{figure*}

A transportation model, whether it simulates traffic or supports a planning study, can be no more accurate than the road network behind it. It takes the network as given: building it is the model's step 0 (Fig.~\ref{fig:stepzero}). For microscopic simulation, signal timing and intersection capacity analysis, that network must be lane-level: how many lanes each approach of an intersection has, which movements each lane serves and which exit lane each movement enters. Open formats such as the General Modeling Network Specification (GMNS) \citep{gmns,berg2022gmns} store exactly this, and open tools carry it to models of every resolution (Section~\ref{sec:related}), yet the lanes are still coded by hand. In an FHWA review of practice built on interviews with 19 agencies and companies \citep{zhou2021mrm}, analysts take a network from GIS files and supply the junction and movement data manually. Building a midsize network for integrated mesoscopic and microscopic modelling takes a team months, and an update of the base year takes a major effort again. Without a common data format, data must be converted for each tool, which invites errors. Agencies want pilot projects and an analysis shown to be cost-effective before it begins, and the review calls for a tool that converts a link-based network into a lane-based one. Open map data do not close the gap: on our test intersections, OpenStreetMap's lane tags cover 29 of the 44 approaches, and read as tagged they give the correct lane layout for 13 of those 29.

Vision-language models (VLMs) can read road imagery, so it is tempting to ask one for the layout directly. The difficulty lies less in recognising an arrow than in keeping the geometry consistent: the same lane must be found in a satellite crop, in a street-level photo looking back at traffic (where the driver's left appears on the right), in photos facing the junction from different distances, and in an ID-labelled network. A curbside strip may be parking upstream and a right-turn lane at the stop bar, and a lane counted at one step must stay the same lane when its turns are read at the next. Asked once with a satellite crop, Claude Opus 5.5 reads 54.5\% of approach layouts exactly; with street views added on the same sheet it reads 50.0\%, and GPT-6.1 Sol drops from 63.6\% to 49.2\%. More views on one sheet do not help. Such failures of consistency across viewpoints, fragments and reasoning steps are what this paper addresses.

A coding agent is checked by tests and a game agent by its score \citep{yang2024sweagent,hu2026lmgame}. A road network gives its signal only when a model run on it meets the counts, and a microscopic model of a large network takes a very long time to calibrate and run \citep{zhou2021mrm}. We therefore build a harness, PAVE (Perceive, Anchor, Verify, Edit), in which checks catch answers that contradict a measurement, a second reading or a traffic rule. Its principle is to measure in code and judge with the model: traffic engineering knowledge enters as general rules and checks, and the model makes only the open judgements, and each must name the network element or image coordinate it refers to, so that code can check it and a lane keeps one identity across views and steps.

PAVE handles two tasks. Network understanding, a problem of spatial perception, recovers the lanes of each approach, their turns and their exit lanes. Scenario editing, a problem of spatial construction, changes a facility to a planning brief. Both write into the GMNS tables a transportation model runs on, keyed to the network's links. With PAVE, the two models read 90.9\% and 85.6\% of the 44 approach layouts exactly, averaged over three runs, against at most 63.6\% with a single request. PAVE takes 18--19 requests per intersection; reading whole subareas, re-reading only the approaches a new base year changes, and sending an approach back when the counts reject it are left to future work.

We contribute a harness that anchors a VLM's judgements about a road network to measured geometry and network IDs, so that code can check them across views and steps. We apply it to network understanding from satellite and street-level imagery, evaluated against human labels for two VLMs with and without the harness, and trace the gain to the steps that produce it. We apply it also to scenario editing, which turns a planning brief into a checked edit of an agency network, a build alternative that can be simulated and compared with the no-build network. Last, we analyse where geometric consistency breaks and which step catches each failure.

\section{Related Work}
\label{sec:related}
Benchmarks show that VLMs struggle with spatial relations, depth and correspondence between images \citep{fu2024blink}, and that consistency breaks when the same scene is seen from several viewpoints \citep{yeh2026allangles}, including urban scenes that combine satellite and street-level views \citep{zhou2025urbench}. Training on generated spatial questions improves such reasoning \citep{chen2024spatialvlm}; PAVE leaves the model as it is and moves the measuring into code. Unlike cross-view geo-localisation \citep{zhu2021vigor}, PAVE starts from each photo's recorded pose; its problem is matching individual lanes across views.

A harness, the code that decides what a fixed model sees and which answers stand, shapes how well coding and game agents do and can itself be optimised \citep{xia2025agentless,han2026vista,lee2026metaharness}. Program-based visual reasoning composes vision tools in code \citep{gupta2023visprog,suris2023vipergpt}, visual prompting adds marks the model can refer to \citep{yang2023som}, and a model can search an image for small details \citep{wu2024vstar} or draw auxiliary lines on it \citep{hu2024sketchpad}; PAVE's detail crops, rulers and stations serve these ends, cut and drawn by code. Without an external signal, self-correction does not improve reasoning \citep{huang2024cannot}; it helps when tools or verifiers supply the signal \citep{gou2024critic}; PAVE's verifiers are geometric measurements and traffic rules. Self-consistency takes the majority of one question asked several times \citep{wang2023selfconsistency}; PAVE asks two different questions of each approach, a trace and a blind reading, and changes its lanes only when the second agrees.

Lane graphs are extracted from aerial imagery \citep{he2022lanelevel}, and learned models infer lane counts and turn types at signalised intersections from aerial imagery and OpenStreetMap \citep{nilsson2026predicting}. In driving, lane topology is inferred from on-board cameras \citep{wang2023openlanev2}. PAVE instead uses a general VLM without task-specific training, adds street-level panoramas, and is not compared with these trained models.

Among road-scene benchmarks for multimodal models, RoadBench \citep{zhang2026roadbench} is the closest work. It asks multimodal models for the lane count and the turns of each lane on a road segment, from a satellite crop, a street photo or both, and finds that counting is weak, that turn designation seldom beats a table of common traffic rules, and that most models score lower on its cross-view tasks, which use different photos, than on the street photo alone. Each case is one segment and one in-vehicle photo, without exit lanes or network links; PAVE reads whole approaches from several panoramas, keyed to network IDs. Our single requests, too, count poorly and gain nothing from added street views.

In practice, a lane-based network keeps macroscopic, mesoscopic and microscopic models consistent, since each link maps to lanes and each lane to cells \citep{zhou2021mrm}. On GMNS, osm2gmns builds links from OpenStreetMap \citep{lu2023virtual}, net2cell expands them into lanes and cells for simulation \citep{net2cell} and DTALite assigns demand to them \citep{zhou2014dtalite}; PAVE supplies the lanes and turns they take as input. Learning-based signal control, in contrast, is mostly tested on synthetic grids \citep{wei2019tscsurvey}.

LLM agents build and edit SUMO scenarios \citep{li2024chatsumo}, call network tools such as osm2gmns \citep{da2024openti}, or write GMNS files from text or images \citep{chen2024ngai}, and validate with simulation results or tool errors. PAVE edits an agency network to a planning brief through design templates \citep{cunningham2021ddi,messer1991spui}, checks the result with traffic engineering rules and compares it with the agency's own design. To our knowledge, no prior work combines reading the lane layouts of whole intersection approaches keyed to network links, checking model answers with traffic knowledge, and editing a real network to a brief compared with an agency design.

\section{The PAVE Harness}
\label{sec:harness}

\subsection{Problem Setting}
PAVE works on a macroscopic road network in GMNS, of which network understanding uses only the geometry and IDs of nodes and links: the lane counts and turning movements stored in the network are never shown to the model. A scenario edit is shown the tables its brief needs (Appendix~\ref{app:edit}). Alongside the network it takes georeferenced imagery: a satellite image and street-level panoramas, each with its camera pose and capture date. The output keeps what is observed, in the network and the imagery metadata, apart from what is inferred: a proposed change to the network with an evidence record that ties every changed element to what the model saw, for the modeller who keeps the network to accept or reject. Each arm of an intersection has an approach, whose lanes are counted at the stop bar and listed from the driver's left, and an exit.

\begin{figure*}[t]
\centering
\includegraphics[width=\textwidth]{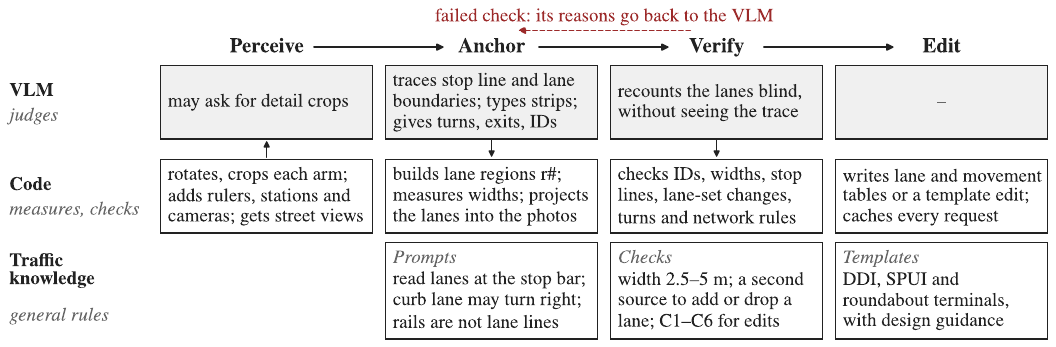}
\caption{The PAVE harness: for each step, what the VLM judges (top), what code does (middle) and the traffic knowledge that enters as general rules (bottom). A failed check sends its reasons back to the VLM.}
\label{fig:harness}
\end{figure*}

Code measures widths, distances, headings, camera projections, IDs and coordinates; the model judges only what measurement cannot settle, such as whether a strip of pavement is a lane, which movements a lane serves or which link is the off-ramp of an interchange. The four steps of the harness follow from this division (Fig.~\ref{fig:harness}).

\subsection{Traffic Knowledge in Prompts, Checks and Templates}
PAVE adds, in three places, the traffic engineering knowledge a person would apply at any intersection. Each rule states general traffic knowledge and names no site, lane count or model. The rules assume right-hand traffic and US markings \citep{fhwa2023mutcd}.

\emph{Prompts} state general rules for reading lanes. A strip's role is judged where it meets the stop bar, because a parking lane upstream often becomes a turn lane at the junction. A lane is recognised by curbs, moving traffic and lane lines that continue to the stop bar, not by the colour of the pavement. The rightmost lane may turn right unless a marking or a fixed sign rules it out. The rails of a streetcar laid in a lane are not lane lines. Curbside parking ends short of a junction, so a lane-wide strip beyond a bicycle lane is normally a right-turn lane at the stop bar. The network records the permanent layout, so temporary traffic control is reported separately and changes neither a lane's existence nor its turns.

\emph{Checks} encode consequences of the same knowledge. A traced motor lane narrower than 2.5\,m or wider than 5\,m is sent back for a second look, but widths never decide a strip's role on their own, since true curbside lanes as narrow as 2.2\,m exist. If a blind reading of the approach, made without the trace, counts more motor lanes than the trace, the arm is traced again. A lane judged not to exist carries no turns, and a lane that allows a turn must connect to a lane of that exit. For editing, each interchange form has required and prohibited movements, an arterial may not cross a freeway at grade, and nothing outside the facility, the mainline included, may change.

\emph{Templates} hold the geometry and movement pattern of each design an edit can build, with its own checks and notes from design guidance. The library holds three interchange designs so far, the diverging diamond (DDI), the single-point urban interchange (SPUI) and roundabout ramp terminals \citep{cunningham2021ddi,messer1991spui}.

\subsection{Perceive}
Code renders every view the model sees at a known scale and orientation. For an intersection approach, the satellite crop is rotated so that inbound traffic drives up the image. It carries pixel rulers and lines at fixed distances from the junction (stations), so that every answer can be given as coordinates. Street-level views come from three poses: a near view looking back at approaching traffic, in which the driver's left appears on the right of the photo, and mid and far views facing the junction. Each camera position is marked on the satellite crop. The model may ask for up to three enlarged crops of any region, cut by code from the source images. PAVE uses no detector or segmentation model: whether a strip is a lane at the stop bar is not an object class a detector would find.

\subsection{Anchor}
Every answer is phrased in terms that code can check. A trace gives the stop line and each longitudinal boundary as coordinates on the ruler frame, and names what lies between each pair of boundaries: a motor lane, a bicycle lane, parking, a median and so on. Later answers refer to the lane regions that code derived from the trace ($r_1, r_2, \dots$) and to the lanes of each exit. An edit names the link and node IDs bound to each part of a facility. Code turns these anchors into measurements, and the model then chooses among candidates that code has measured, rather than producing geometry.

\subsection{Verify}
Checks fall into four groups. The first checks format and IDs: every answer is validated against its schema and against the IDs and regions it was shown; for editing, gates (checks on the model's answer before anything is built) also check the structure the brief requires. The second checks geometry: lane widths, boundaries that must not cross, and stop lines. The third compares views: a blind reading of the satellite crop and street views, which never sees the trace, is compared with the traced lane count, and traced lanes are projected into the street views. The fourth checks the network: for an interchange, checks C1--C6 cover table integrity, that no trip possible before the edit is lost and the rules above, followed by each template's own checks. The other scenario types have checks of their own (Appendix~\ref{app:edit}). In understanding, a failed check returns its reasons to the model, within a fixed repair budget; in editing, only a failed gate does, and the network checks are reported with the edit.

\subsection{Edit}
The checked result is written out against network IDs: lane layouts and lane pairs, or network tables and a table of changes. Every model request is answered by a fresh model instance that sees only that request, and every answer is cached under a signature of its images, prompt and model. A run can therefore be replayed exactly from its cache, and each element of a result can be traced to the answer and the views that produced it.

\section{Network Understanding}
\label{sec:understanding}
\begin{figure}[t]
\centering
\includegraphics[width=\columnwidth]{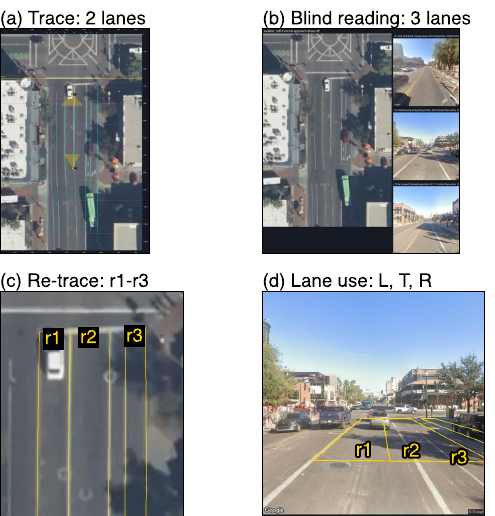}
\caption{One approach through the harness: the re-trace (c) finds the lane that the first trace (a) missed and the blind reading (b) counted; (d) projects the lanes into a photo facing the junction, as the model saw it. Cyan: traced lanes. Satellite imagery: \textcopyright\,Mapbox, \textcopyright\,OpenStreetMap, \textcopyright\,Maxar; street-level imagery: \textcopyright\,Google.}
\label{fig:understanding}
\end{figure}

Understanding an intersection means recovering, for every approach, how many lanes reach the stop bar, their order from the driver's left, the movements each lane allows and the exit lane each movement enters, with every item tied to a network link and its lane regions. PAVE reaches this through a short chain of requests (Fig.~\ref{fig:understanding}). In the steps of Section~\ref{sec:harness}, code perceives for every request, the trace and lane use anchor, the blind reading, re-trace and lane-set rules verify, and the tables are the edit. Crop size and placement follow from link geometry alone, and no lane count from the network (except to choose where street views are requested), no reference label and no answer from another run enters a prompt or decides which requests are made.

The trace takes one request per road arm. On the rotated satellite crop, the model marks the stop line, the start of the exit and every longitudinal boundary of both carriageways, and names what lies between each pair of boundaries. Code turns the boundaries into lane regions and measures their widths at fixed distances from the junction.

The blind reading takes one request per approach, independent of the trace. The model sees the unannotated satellite crop and the approach's street views, reports the lane layout at the stop bar, and says whether its lane count is settled. Never seeing the trace, it cannot copy the trace's mistakes, so a disagreement marks an approach to look at again.

An arm is traced again only where a settled blind reading sees more motor lanes than the trace or a measured issue applies (a motor lane outside 2.5--5\,m or marked low-confidence, a bicycle strip at least 2.5\,m wide between two lanes). The trace is shown back with these issues and its boundaries projected into the street views, and the model corrects it or keeps it.

The lane-use step takes one request per approach. The traced lanes are outlined on the satellite crop and projected into the street views, and the prompt gives the traced cross-section: bicycle lanes, parking and medians between the lanes. For each lane, the model judges whether it is a motor lane at the stop bar and which turns it allows, and for each turn it gives the exit and the lane pairs.

Changes to the lane set need a second source. The lane-use answer only proposes them, and other evidence decides. A lane reported at the stop bar without an outline of its own is adopted only if a new trace of the arm finds it. A traced lane judged not to exist is dropped only when the blind reading is settled and counts no more lanes than would remain, and a lane at least 2.5\,m wide is asked about once more before it is dropped. Two further rules re-ask about left turns (Appendix~\ref{app:understanding}).

For a transportation model, every approach and exit carries its link ID, so the result maps onto GMNS: lane counts onto links, lane indices and widths onto the lane table, and each turn's lane pairs onto the inbound and outbound lanes of its movement record, from which simulation reads capacity, turn restrictions and lane-to-lane connections (Fig.~\ref{fig:app-results}) and net2cell builds lane-by-lane cells \citep{net2cell}. The records are proposals for a modeller to review, not changes to the network: the modeller's work shifts from coding each approach to reviewing what PAVE proposes. PAVE does not measure turn-lane lengths.

\section{Scenario Editing}
\label{sec:edit}
Scenario editing runs the same four steps on a larger unit: code renders one image, the binding request anchors, gates and checks verify, and the build is the edit. Before a diamond interchange can become a diverging diamond, the model has to say which existing links are the ramps, the ramp terminals and the crossing road; this binding is network understanding for a facility.

In the binding request, the model sees one rendered image, a satellite panel next to a drawing of the network that labels every link and node ID on a grid, together with the planning brief and the link and node tables (Appendix~\ref{app:edit} lists what each scenario shows). Each scenario type has its own query, with a shared system prompt and outline (Appendix~\ref{app:setup}). The model returns a single answer: the facility class, the IDs bound to each part of the facility, the operations to apply, its evidence and what it could not establish.

Gates (P0--P8 for an interchange) check the answer; a failed answer gets one repair request listing its defects. After the build, the network checks (C1--C6 for an interchange), the template's own checks and the design guidance test the result. The checks cover topology and table integrity; the capacity and signal control of new links come from the template, for a modeller to confirm.

Code then applies the operations, builds the design from its template and writes the after-network in GMNS with a table of changes.

\section{Experiments}
\label{sec:experiments}

\subsection{Setup}
Network understanding is evaluated on 11 signalised intersections in Tempe, Arizona, with 44 approaches and 43 exits (Appendix~\ref{app:scoring} describes them). Every approach's lanes and turns were labelled by hand from the imagery. Scenario editing uses three scenarios from the Arizona DOT, each a no-build network and a brief: S1 adds a lane on US-60 (the model names the links to widen), S2 converts the SR~347 / I-10 diamond interchange into a DDI (it binds six roles at each of the two ramp terminals) and S3 adds passing lanes on US-60 (it gives the start and end of each). These are typical uses of dynamic traffic assignment \citep{zhou2021mrm}, on an agency's network rather than a synthetic grid. The agency's build networks, which the model never sees, are the reference design.

We test Claude Opus 5.5 and GPT-6.1 Sol. As a single request, each model answers one request per intersection, with one image and no tools, in two variants: the satellite crop alone, and the satellite crop with the approaches' street views on one sheet; each variant is repeated three times. With PAVE, each model runs the full harness of Section~\ref{sec:harness} three times, all with the same code version, making 196--213 requests per run against 11 per repeat for a single request (Table~\ref{tab:requests} in Appendix~\ref{app:understanding}). The single request also lacks PAVE's reading rules and shows each view at a lower resolution, so the comparison measures the whole harness. For editing, each model answers each scenario three times under three conditions (Appendix~\ref{app:edit}): PAVE; no harness, where the same model works as a general agent that gets the same brief, image and No-Build network files and writes and runs its own code, with no gates, repair, template or checks; and PAVE without gates, where the first answer of each PAVE run goes straight to the template.

Each approach is scored against the labels on the lane count; the layout, meaning every lane's set of turns in order from the driver's left (the main metric); the set of turns the approach allows; and the number of lanes serving each turn. U-turns are not scored. For differences we report a 95\% bootstrap interval that resamples intersections; permutation tests are in Appendix~\ref{app:scoring}. An edit is scored on the network checks it passes and on its agreement with the agency's design.

\begin{table}[t]
\centering
\small
\begin{tabular}{@{}lccccc@{}}
\toprule
 & & & Turn & Turn & \\
Condition & Count & Layout & set & lanes & Exits \\
\midrule
\multicolumn{6}{@{}l}{\emph{Claude Opus 5.5}} \\
Satellite & 64.4 & 54.5 & 91.7 & 75.8 & -- \\
Sat.\ + street & 67.4 & 50.0 & 90.9 & 66.7 & -- \\
Blind reading & 86.4 & 48.5 & 62.1 & 50.8 & -- \\
PAVE & \textbf{90.9} & \textbf{90.9} & \textbf{100.0} & \textbf{99.2} & \textbf{96.1} \\
\midrule
\multicolumn{6}{@{}l}{\emph{GPT-6.1 Sol}} \\
Satellite & 67.4 & 63.6 & 93.9 & 81.1 & -- \\
Sat.\ + street & 58.3 & 49.2 & 91.7 & 75.8 & -- \\
Blind reading & 74.2 & 48.5 & 66.7 & 51.5 & -- \\
PAVE & \textbf{89.4} & \textbf{85.6} & \textbf{97.0} & \textbf{90.2} & \textbf{93.8} \\
\bottomrule
\end{tabular}
\caption{Network understanding on 44 approaches (exits: 43), percent correct. Satellite and Sat.\ + street: one request per intersection. Blind reading: PAVE's request per approach without the trace, lanes without arrows scored by a fixed rule (Appendix~\ref{app:scoring}). Means of three repeats or runs. Exits: lane count of each exit, which only PAVE reports. Bold: best in each column for each model.}
\label{tab:understanding}
\end{table}

\begin{figure*}[t]
\centering
\includegraphics[width=\textwidth]{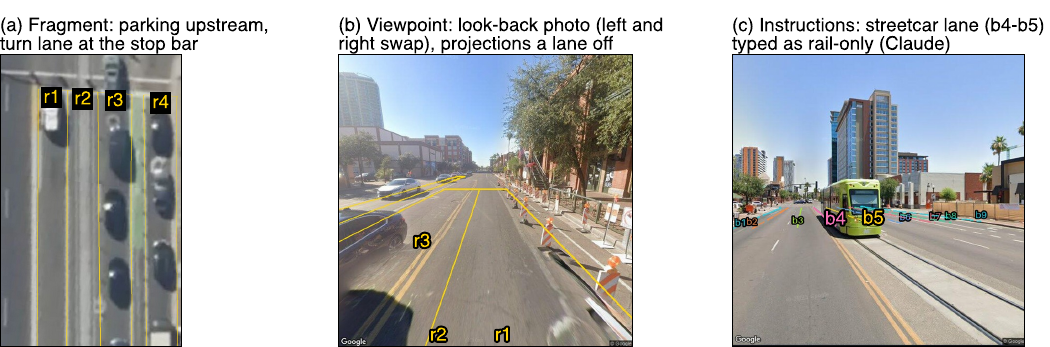}
\caption{Where geometric consistency breaks: one example each of a fragment, a viewpoint and an instruction error. Satellite imagery: \textcopyright\,Mapbox, \textcopyright\,OpenStreetMap, \textcopyright\,Maxar; street-level imagery: \textcopyright\,Google.}
\label{fig:failures}
\end{figure*}

\subsection{Network Understanding}
\label{sec:accuracy}
With PAVE, Claude's three runs read 88.6--93.2\% of the 44 layouts exactly and GPT's 77.3--90.9\% (Table~\ref{tab:understanding}). Against the same model as a single request, Claude gains 36.4 percentage points over the satellite variant (95\% interval 15.2--54.6) and 40.9 points over the variant with street views (18.9--59.8); GPT gains 22.0 points (10.6--34.1) and 36.4 points (24.2--49.2). PAVE's blind reading alone already counts the lanes right on 86.4\% of the approaches with Claude and 74.2\% with GPT, but its layouts are no better than a single request's (48.5\% with either model), since it reports only the turns that arrows show; PAVE is 42.4 points (25.0--59.8) and 37.1 points (20.5--53.8) above it. On the 29 approaches that OpenStreetMap tags, its lane tags give 44.8\% correct layouts (65.5\% if the rightmost lane is also allowed to turn right), the single requests 48.3--64.4\% on average, and PAVE 93.1\% with Claude and 86.2\% with GPT.

We group the failures by the three axes of geometric consistency, and add a fourth group for the instructions (Fig.~\ref{fig:failures}); Table~\ref{tab:failures} in Appendix~\ref{app:scoring} counts the errors by kind and the step that catches them. Most single-request errors miss a lane. Across viewpoints, adding street views to the satellite crop does not raise the single request's layout accuracy (Claude 54.5\% and 50.0\%, a difference of 4.5 points, $-$7.6 to 18.2; GPT 63.6\% and 49.2\%, 14.4 points, 3.8 to 25.8). It adds wrong turns on correctly counted approaches (from 3.3 to 7.3 for Claude and from 1.0 to 3.3 for GPT, means of three repeats), while a whole layout read mirrored stays rare. Each view gets fewer pixels on one sheet, so this does not isolate the views. The look-back photo mirrors left and right, and camera positions can be off by several metres, so lanes are easily matched to the wrong strip. In each PAVE run the lane-use answers note a projection offset, often by a lane, on at least 14--17 (Claude) and 17--24 (GPT) of the 44 approaches, and there the model, not code, matches the lanes across the views by their lane lines. Fragments of a lane mislead too: a curbside strip that is parking upstream may be a turn lane at the stop bar, and a strip with special paving may be a turn lane or a median. Most remaining errors are of this kind. Of Claude's 12 wrong layouts over the three runs, 10 are a curbside strip kept as a lane or left out; of GPT's 19, 14 are a lane left out, mostly typed as a median, parking or bicycle strip, and five are wrong turns. Across reasoning steps, of the lane counts the trace gets wrong (7.3 per run for Claude, 9.7 for GPT), the blind reading flags 1.3 and 3.0, a measured width or low confidence 3.7 and 4.0 more, and 2.3 and 2.7 go unflagged. The model must also apply each rule as intended. The prompts count a lane shared with a streetcar, yet in every run Claude's trace typed such a lane, which also takes left turns, as rail-only (Fig.~\ref{fig:failures}c); a later step restored it in one run, and the other two runs give Claude's two remaining errors.

\begin{table}[t]
\centering
\small
\begin{tabular}{@{}lcccc@{}}
\toprule
 & \multicolumn{2}{c}{Claude Opus 5.5} & \multicolumn{2}{c}{GPT-6.1 Sol} \\
\cmidrule(lr){2-3}\cmidrule(lr){4-5}
Stage & Count & Layout & Count & Layout \\
\midrule
Single request, satellite & 64.4 & 54.5 & 67.4 & 63.6 \\
Single request, + street & 67.4 & 50.0 & 58.3 & 49.2 \\
\midrule
After the trace & 83.3 & -- & 78.0 & -- \\
After the re-trace & 82.6 & -- & 83.3 & -- \\
First lane use & -- & 83.3 & -- & 78.0 \\
After detail crops & -- & 83.3 & -- & 78.8 \\
Final & \textbf{90.9} & \textbf{90.9} & \textbf{89.4} & \textbf{85.6} \\
\bottomrule
\end{tabular}
\caption{Percent of the 44 approaches correct after each stage of PAVE, mean of each model's three runs, with the single requests of Table~\ref{tab:understanding} for comparison. Layouts are first scored at the first lane use; a lane that the lane-use answer judges not to exist is dropped before scoring. Bold: best in each column.}
\label{tab:stages}
\end{table}
Most of the gain comes from reading one approach at a time on traced lanes. Table~\ref{tab:stages} follows the runs through the harness: at the first lane-use answer, with each exit's direction and the turn rules, the layouts are already 83.3\% right with Claude and 78.0\% with GPT, against 49.2--63.6\% for a single request. The re-trace guards against a poor trace: after GPT's first trace (68.2\% of counts right) it fixes nine counts and breaks one, while in the other five runs it breaks about as many as it fixes (Appendix~\ref{app:scoring}). The rules for changing the lane set act three to nine times per run and fix two to five counts and two to five layouts, breaking one count once and no layout. The detail crops, a second look at small arrows and signs, change a lane in 11 of their 194 rounds, eight of them inside the re-asks of the lane-set rules.

Not every piece of knowledge pays off in accuracy. The rule that the rightmost lane may turn right agrees with the labels on all 44 approaches in every Claude run and on 42--43 in GPT's, but the single requests already agree on 41--43, so the gain does not come from it. Nor do rules alone give the turns: given the labelled lane counts, a table of common rules gets 19 of the 44 layouts right (Appendix~\ref{app:scoring}).

\subsection{Scenario Editing}
\begin{table}[t]
\centering
\small
\begin{tabular}{@{}lccc@{}}
\toprule
Condition & S1 (6 checks) & S2 (9 checks) & S3 (10 checks) \\
\midrule
\multicolumn{4}{@{}l}{\emph{Claude Opus 5.5}} \\
No harness & 77.8 & 77.8 & 83.3 \\
PAVE & \textbf{100.0} & \textbf{100.0} & \textbf{100.0} \\
\midrule
\multicolumn{4}{@{}l}{\emph{GPT-6.1 Sol}} \\
No harness & 83.3 & 88.9 & 80.0 \\
PAVE & \textbf{100.0} & \textbf{100.0} & \textbf{93.3} \\
\bottomrule
\end{tabular}
\caption{Scenario editing: percent of the checks passed over the three runs. No harness: the same model as a general agent writing its own code. Bold: the better value for each model.}
\label{tab:edit}
\end{table}

The conditions bind the facility alike. In S1 both widen the agency's links, in S2 a short script reading only node and facility types binds all twelve roles as PAVE does (Appendix~\ref{app:edit}), and in S3 both place the passing lanes within 0.1\,km of each other. They differ in building and checking the edit.

Building from templates is what makes the edits pass the checks. With PAVE, 16 of the 18 runs pass every check, and no agent run does (Table~\ref{tab:edit}). Without gates, every run passes the same checks: the gates fired only on GPT's three first answers for S3 (P6), and two of these, repaired or not, leave trucks out of the general lane and fail C5, which the gates do not cover. A failed check counts as a convention of the checks, not an error, when the network still works as designed (Appendix~\ref{app:edit}). Setting these aside, 7 of the 18 agent runs contain a design or coding error (Table~\ref{tab:agentfail}), against 2 of the 18 PAVE runs; six of the seven leave the new geometry inconsistent with the network around it, such as lanes that dead-end 2.2\,m from a junction node. Errors cluster in S3, with five of the agents' seven and both of PAVE's, so the counts describe three scenarios, not 18 independent trials.

A brief allows more than one valid design, so the agency's design is a reference, not a target, and agreement with it is similar with and without the harness. In S2 the crossovers lie 2.9--3.8\,m from the agency's crossing points with PAVE, less than a lane width, and 2.6--12.3\,m without it. In S3 both conditions cover the length of the agency's westbound passing lane, and 40\% (PAVE) or 36\% of its eastbound one, which ends 0.9\,km short of where the brief has it end. In S1 PAVE also scales a fourth period that the agency's design leaves unchanged.

\section{Discussion and Limitations}
\label{sec:limits}
Within the harness, the gain appears at two steps. Asking about one approach at a time, with views made for it, gets the lane counts right (Section~\ref{sec:accuracy}), though the single request also lacks the reading rules. Layouts come right at the lane-use step, which reads turns on traced lanes, given each exit's direction and the turn rules; our conditions do not separate these three inputs. A per-approach single request with the same rules and exit directions but no trace would.

Three practices of the harness do not depend on roads: ask about one unit at a time, with views made for it; have the model refer to regions and IDs that code drew and projected, rather than produce geometry; and let a proposal to change the structure stand only when a different question or a measurement agrees. They need measurable geometry and a structure that can carry IDs; we tested them only on roads.

For practice, OpenStreetMap is the baseline: of the 44 approaches, a network built from its lane tags would need 31 layouts coded or corrected (25 if its rightmost lanes may also turn right), against three to five per run with Claude and four to ten with GPT. PAVE's own marks of doubt cannot yet tell a modeller which approaches to check: over the three runs they fall on 20 approaches holding 6 of Claude's 12 errors, and on 13 approaches holding 3 of GPT's 19, so every approach still needs a look. PAVE makes 18 times as many requests as a single request (Appendix~\ref{app:understanding}). Running the model on its tables could automate the return from the counts: a counted turn that no lane serves rejects a layout outright, while a lane too many or too few shows in counts only near capacity (Appendix~\ref{app:model}). We have not run this check. The evaluation is limited to 11 intersections in one city, and PAVE's correct layouts vary between runs by up to two approaches with Claude and six with GPT, so differences of a few approaches lie within this spread. The blind reading comes from the same model as the trace and can share its errors. The lane pairs to exit lanes are not scored. Some street views are years old, and some show temporary traffic control. Editing covers three scenarios in one state. The two tasks meet in the GMNS tables but are tested apart: every edit takes its lane counts from the agency's network.

\section{Conclusion}
PAVE lets a VLM read and edit a road network by dividing the work: code measures, traffic knowledge enters as rules and checks, and the model judges only what measurement cannot settle, anchored to the network. On 44 approaches PAVE raises the share of exactly correct lane layouts from about half to 91\% for Claude and from about two thirds to 86\% for GPT. Most of this gain comes at the first lane-use answer, which reads one approach at a time on traced lanes; the second reading and the checks fix a third to a half of the layouts still wrong. In scenario editing the model names the parts of a facility and code builds the design from a template: 16 of 18 edits pass every network check, against none without the harness, and 2 contain a design or coding error, against 7, while agreement with the agency's design is about the same either way. Where OpenStreetMap's lane tags leave 31 of 44 layouts to correct, PAVE leaves three to ten per run, a step toward the link-to-lane conversion that practitioners ask for.

\section*{Acknowledgments}
The authors acknowledge the OpenStreetMap community for providing the roadway data used to construct the baseline General Modeling Network Specification network. Satellite imagery is from Mapbox, street-level imagery from Google Street View, and map data \copyright{} OpenStreetMap contributors. This research was supported by the National Science Foundation (Award No.~2303748) and by the U.S. Department of Transportation through the Advancing Community-Centric Transportation Systems (ACTS) Center under Federal Grant No.~693JK42550003.

\bibliography{refs}

\begin{thebibliography}{40}
\providecommand{\natexlab}[1]{#1}

\bibitem[{{Arizona Department of Transportation}(2025)}]{adot2021rdg}
{Arizona Department of Transportation}. 2025.
\newblock \emph{Roadway Design Guidelines}.
\newblock Arizona Department of Transportation.
\newblock 2021 edition with {January} 2025 revisions.

\bibitem[{Berg, Smith, and Zhou(2022)}]{berg2022gmns}
Berg, I.; Smith, S.; and Zhou, X.~S. 2022.
\newblock {GMNS}: A Specification for Sharing Routable Road Networks.
\newblock In \emph{Transportation Research Board 101st Annual Meeting},
  TRBAM-22-02127. Washington, DC.

\bibitem[{Chen et~al.(2024)Chen, Xu, Kirmani, Ichter, Sadigh, Guibas, and
  Xia}]{chen2024spatialvlm}
Chen, B.; Xu, Z.; Kirmani, S.; Ichter, B.; Sadigh, D.; Guibas, L.; and Xia, F.
  2024.
\newblock {SpatialVLM}: Endowing Vision-Language Models with Spatial Reasoning
  Capabilities.
\newblock In \emph{Proceedings of the IEEE/CVF Conference on Computer Vision
  and Pattern Recognition (CVPR)}, 14455--14465.

\bibitem[{Chen et~al.(2026)Chen, Xu, Cao, Xu, Zhang, Zhang, Zhang, and
  Yu}]{chen2024ngai}
Chen, J.; Xu, W.; Cao, H.; Xu, Z.; Zhang, Y.; Zhang, S.; Zhang, Z.; and Yu, B.
  2026.
\newblock Network Generation Artificial Intelligence: Multimodal Road Network
  Generation Based on Large Language Models.
\newblock \emph{Transportation Research Record}, 2680(1): 432--451.

\bibitem[{Cunningham et~al.(2021)Cunningham, Chase, Deng, Carnes, Pyo, Jenior,
  Schroeder, Ray, Urbanik, Knudsen, Rodegerdts, Warchol, and
  Tanaka}]{cunningham2021ddi}
Cunningham, C.; Chase, T.; Deng, Y.; Carnes, C.; Pyo, K.; Jenior, P.;
  Schroeder, B.; Ray, B.; Urbanik, T.; Knudsen, J.; Rodegerdts, L.; Warchol,
  S.; and Tanaka, A. 2021.
\newblock Diverging Diamond Interchange Informational Guide, Second Edition.
\newblock NCHRP Research Report 959, Transportation Research Board, National
  Academies of Sciences, Engineering, and Medicine.

\bibitem[{Da et~al.(2024)Da, Liou, Chen, Zhou, Luo, Yang, and
  Wei}]{da2024openti}
Da, L.; Liou, K.; Chen, T.; Zhou, X.; Luo, X.; Yang, Y.; and Wei, H. 2024.
\newblock Open-{TI}: Open Traffic Intelligence with Augmented Language Model.
\newblock \emph{International Journal of Machine Learning and Cybernetics},
  15(10): 4761--4786.

\bibitem[{{Federal Highway Administration}(2023)}]{fhwa2023mutcd}
{Federal Highway Administration}. 2023.
\newblock \emph{Manual on Uniform Traffic Control Devices for Streets and
  Highways}.
\newblock U.S. Department of Transportation, 11th edition.

\bibitem[{Fu et~al.(2024)Fu, Hu, Li, Feng, Wang, Lin, Roth, Smith, Ma, and
  Krishna}]{fu2024blink}
Fu, X.; Hu, Y.; Li, B.; Feng, Y.; Wang, H.; Lin, X.; Roth, D.; Smith, N.~A.;
  Ma, W.-C.; and Krishna, R. 2024.
\newblock {BLINK}: Multimodal Large Language Models Can See but Not Perceive.
\newblock In \emph{Computer Vision -- ECCV 2024}, Lecture Notes in Computer
  Science, 148--166. Springer.

\bibitem[{Gou et~al.(2024)Gou, Shao, Gong, Shen, Yang, Duan, and
  Chen}]{gou2024critic}
Gou, Z.; Shao, Z.; Gong, Y.; Shen, Y.; Yang, Y.; Duan, N.; and Chen, W. 2024.
\newblock {CRITIC}: Large Language Models Can Self-Correct with
  Tool-Interactive Critiquing.
\newblock In \emph{International Conference on Learning Representations
  (ICLR)}.

\bibitem[{Gupta and Kembhavi(2023)}]{gupta2023visprog}
Gupta, T.; and Kembhavi, A. 2023.
\newblock Visual Programming: Compositional Visual Reasoning Without Training.
\newblock In \emph{Proceedings of the IEEE/CVF Conference on Computer Vision
  and Pattern Recognition (CVPR)}, 14953--14962.

\bibitem[{Han et~al.(2026)Han, Hu, Qiu, Wu, and He}]{han2026vista}
Han, Q.; Hu, K.; Qiu, L.; Wu, C.; and He, K. 2026.
\newblock {VISTA}: A Visual Harness for Reasoning in an Interactive World.
\newblock arXiv:2610.02200.

\bibitem[{He and Balakrishnan(2022)}]{he2022lanelevel}
He, S.; and Balakrishnan, H. 2022.
\newblock Lane-Level Street Map Extraction From Aerial Imagery.
\newblock In \emph{Proceedings of the IEEE/CVF Winter Conference on
  Applications of Computer Vision (WACV)}, 2080--2089.

\bibitem[{Hu et~al.(2026)Hu, Huo, Zhang, Yu, Xing, Stoica, Rosing, Jin, and
  Zhang}]{hu2026lmgame}
Hu, L.; Huo, M.; Zhang, Y.; Yu, H.; Xing, E.~P.; Stoica, I.; Rosing, T.; Jin,
  H.; and Zhang, H. 2026.
\newblock lmgame-Bench: How Good are {LLMs} at Playing Games?
\newblock In \emph{International Conference on Learning Representations}.

\bibitem[{Hu et~al.(2024)Hu, Shi, Fu, Roth, Ostendorf, Zettlemoyer, Smith, and
  Krishna}]{hu2024sketchpad}
Hu, Y.; Shi, W.; Fu, X.; Roth, D.; Ostendorf, M.; Zettlemoyer, L.; Smith, N.;
  and Krishna, R. 2024.
\newblock Visual Sketchpad: Sketching as a Visual Chain of Thought for
  Multimodal Language Models.
\newblock In \emph{Advances in Neural Information Processing Systems},
  volume~37, 139348--139379.

\bibitem[{Huang et~al.(2024)Huang, Chen, Mishra, Zheng, Yu, Song, and
  Zhou}]{huang2024cannot}
Huang, J.; Chen, X.; Mishra, S.; Zheng, H.~S.; Yu, A.~W.; Song, X.; and Zhou,
  D. 2024.
\newblock Large Language Models Cannot Self-Correct Reasoning Yet.
\newblock In \emph{International Conference on Learning Representations
  (ICLR)}.

\bibitem[{Lee et~al.(2026)Lee, Nair, Zhang, Lee, Khattab, and
  Finn}]{lee2026metaharness}
Lee, Y.; Nair, R.; Zhang, Q.; Lee, K.; Khattab, O.; and Finn, C. 2026.
\newblock Meta-Harness: End-to-End Optimization of Model Harnesses.
\newblock arXiv:2603.28052.

\bibitem[{Li, Azfar, and Ke(2025)}]{li2024chatsumo}
Li, S.; Azfar, T.; and Ke, R. 2025.
\newblock {ChatSUMO}: Large Language Model for Automating Traffic Scenario
  Generation in Simulation of Urban {MObility}.
\newblock \emph{IEEE Transactions on Intelligent Vehicles}, 10(11): 4962--4973.

\bibitem[{Lu, Liu, and Zhou(2026)}]{net2cell}
Lu, J.; Liu, Y.; and Zhou, X. 2026.
\newblock net2cell: A Multi-Resolution Network Builder for {GMNS} Networks.
\newblock \url{https://pypi.org/project/net2cell/}.
\newblock Python package, version 0.1.4.

\bibitem[{Lu and Zhou(2023)}]{lu2023virtual}
Lu, J.; and Zhou, X.~S. 2023.
\newblock Virtual track networks: A hierarchical modeling framework and
  open-source tools for simplified and efficient connected and automated
  mobility ({CAM}) system design based on general modeling network
  specification ({GMNS}).
\newblock \emph{Transportation Research Part C: Emerging Technologies}, 153:
  104223.

\bibitem[{Messer et~al.(1991)Messer, Bonneson, Anderson, and
  McFarland}]{messer1991spui}
Messer, C.~J.; Bonneson, J.~A.; Anderson, S.~D.; and McFarland, W.~F. 1991.
\newblock Single Point Urban Interchange Design and Operations Analysis.
\newblock NCHRP Report 345, Transportation Research Board.

\bibitem[{{National Academies of Sciences, Engineering, and
  Medicine}(2023)}]{nchrp2023roundabouts}
{National Academies of Sciences, Engineering, and Medicine}. 2023.
\newblock Guide for Roundabouts.
\newblock NCHRP Research Report 1043, Transportation Research Board.

\bibitem[{Nilsson and Oucheikh(2026)}]{nilsson2026predicting}
Nilsson, H.; and Oucheikh, R. 2026.
\newblock Predicting the adjacency matrix to construct lane topology of
  signalised intersections with aerial imagery and deep learning.
\newblock \emph{Transportation Research Interdisciplinary Perspectives}, 35:
  101779.

\bibitem[{Sur{\'\i}s, Menon, and Vondrick(2023)}]{suris2023vipergpt}
Sur{\'\i}s, D.; Menon, S.; and Vondrick, C. 2023.
\newblock {ViperGPT}: Visual Inference via Python Execution for Reasoning.
\newblock In \emph{Proceedings of the IEEE/CVF International Conference on
  Computer Vision (ICCV)}, 11888--11898.

\bibitem[{Wang et~al.(2023{\natexlab{a}})Wang, Li, Li, Chen, Sima, Liu, Wang,
  Jia, Wang, Jiang, Wen, Xu, Luo, Yan, Zhang, and Li}]{wang2023openlanev2}
Wang, H.; Li, T.; Li, Y.; Chen, L.; Sima, C.; Liu, Z.; Wang, B.; Jia, P.; Wang,
  Y.; Jiang, S.; Wen, F.; Xu, H.; Luo, P.; Yan, J.; Zhang, W.; and Li, H.
  2023{\natexlab{a}}.
\newblock {OpenLane-V2}: A Topology Reasoning Benchmark for Unified {3D} {HD}
  Mapping.
\newblock In \emph{Advances in Neural Information Processing Systems (Datasets
  and Benchmarks Track)}, volume~36, 18873--18884.

\bibitem[{Wang et~al.(2023{\natexlab{b}})Wang, Wei, Schuurmans, Le, Chi,
  Narang, Chowdhery, and Zhou}]{wang2023selfconsistency}
Wang, X.; Wei, J.; Schuurmans, D.; Le, Q.; Chi, E.; Narang, S.; Chowdhery, A.;
  and Zhou, D. 2023{\natexlab{b}}.
\newblock Self-Consistency Improves Chain of Thought Reasoning in Language
  Models.
\newblock In \emph{International Conference on Learning Representations
  (ICLR)}.

\bibitem[{Wei et~al.(2019)Wei, Zheng, Gayah, and Li}]{wei2019tscsurvey}
Wei, H.; Zheng, G.; Gayah, V.; and Li, Z. 2019.
\newblock A Survey on Traffic Signal Control Methods.
\newblock arXiv:1904.08117.

\bibitem[{Wong and Wong(2003)}]{wong2003lane}
Wong, C.~K.; and Wong, S.~C. 2003.
\newblock Lane-based optimization of signal timings for isolated junctions.
\newblock \emph{Transportation Research Part B: Methodological}, 37(1): 63--84.

\bibitem[{Wu and Xie(2024)}]{wu2024vstar}
Wu, P.; and Xie, S. 2024.
\newblock {V*}: Guided Visual Search as a Core Mechanism in Multimodal {LLMs}.
\newblock In \emph{Proceedings of the IEEE/CVF Conference on Computer Vision
  and Pattern Recognition (CVPR)}, 13084--13094.

\bibitem[{Wu et~al.(2018)Wu, Guo, Xian, and Zhou}]{wu2018hierarchical}
Wu, X.; Guo, J.; Xian, K.; and Zhou, X. 2018.
\newblock Hierarchical travel demand estimation using multiple data sources: A
  forward and backward propagation algorithmic framework on a layered
  computational graph.
\newblock \emph{Transportation Research Part C: Emerging Technologies}, 96:
  321--346.

\bibitem[{Xia et~al.(2025)Xia, Deng, Dunn, and Zhang}]{xia2025agentless}
Xia, C.~S.; Deng, Y.; Dunn, S.; and Zhang, L. 2025.
\newblock Demystifying {LLM}-Based Software Engineering Agents.
\newblock \emph{Proceedings of the ACM on Software Engineering}, 2(FSE):
  801--824.

\bibitem[{Yang et~al.(2024)Yang, Jimenez, Wettig, Lieret, Yao, Narasimhan, and
  Press}]{yang2024sweagent}
Yang, J.; Jimenez, C.~E.; Wettig, A.; Lieret, K.; Yao, S.; Narasimhan, K.; and
  Press, O. 2024.
\newblock {SWE}-agent: Agent-Computer Interfaces Enable Automated Software
  Engineering.
\newblock In \emph{Advances in Neural Information Processing Systems},
  volume~37, 50528--50652.

\bibitem[{Yang et~al.(2023)Yang, Zhang, Li, Zou, Li, and Gao}]{yang2023som}
Yang, J.; Zhang, H.; Li, F.; Zou, X.; Li, C.; and Gao, J. 2023.
\newblock Set-of-Mark Prompting Unleashes Extraordinary Visual Grounding in
  {GPT-4V}.
\newblock arXiv:2310.11441.

\bibitem[{Yeh et~al.(2026)Yeh, Wang, Tong, Cheng, Wang, Chu, Zhai, Chen, Gao,
  and Ma}]{yeh2026allangles}
Yeh, C.-H.; Wang, C.; Tong, S.; Cheng, T.-Y.; Wang, R.; Chu, T.; Zhai, Y.;
  Chen, Y.; Gao, S.; and Ma, Y. 2026.
\newblock Seeing from Another Perspective: Evaluating Multi-View Understanding
  in {MLLMs}.
\newblock In \emph{Proceedings of the AAAI Conference on Artificial
  Intelligence}, volume~40, 12000--12008.

\bibitem[{{Zephyr Foundation}(2020)}]{gmns}
{Zephyr Foundation}. 2020.
\newblock General Modeling Network Specification ({GMNS}).
\newblock \url{https://github.com/zephyr-data-specs/GMNS}.
\newblock Volpe Center/FHWA partnership with the Zephyr Foundation.

\bibitem[{Zhang et~al.(2026)Zhang, Zhang, Feng, Chen, Wang, Liu, Jin, and
  Li}]{zhang2026roadbench}
Zhang, J.; Zhang, X.; Feng, J.; Chen, L.; Wang, J.; Liu, Z.; Jin, D.; and Li,
  Y. 2026.
\newblock {RoadBench}: Benchmarking {MLLMs} on Fine-Grained Spatial
  Understanding and Reasoning under Urban Road Scenarios.
\newblock In \emph{Computer Vision -- ECCV 2026}, Lecture Notes in Computer
  Science, 177--195. Springer.

\bibitem[{Zhou et~al.(2025{\natexlab{a}})Zhou, Yang, Chen, Ye, Bai, Yu, Zhang,
  Lin, He, and Li}]{zhou2025urbench}
Zhou, B.; Yang, H.; Chen, D.; Ye, J.; Bai, T.; Yu, J.; Zhang, S.; Lin, D.; He,
  C.; and Li, W. 2025{\natexlab{a}}.
\newblock {UrBench}: A Comprehensive Benchmark for Evaluating Large Multimodal
  Models in Multi-View Urban Scenarios.
\newblock In \emph{Proceedings of the AAAI Conference on Artificial
  Intelligence}, volume~39, 10707--10715.

\bibitem[{Zhou and Taylor(2014)}]{zhou2014dtalite}
Zhou, X.; and Taylor, J. 2014.
\newblock {DTALite}: A Queue-Based Mesoscopic Traffic Simulator for Fast Model
  Evaluation and Calibration.
\newblock \emph{Cogent Engineering}, 1(1): 961345.

\bibitem[{Zhou, Hadi, and Hale(2021)}]{zhou2021mrm}
Zhou, X.~S.; Hadi, M.; and Hale, D. 2021.
\newblock Multiresolution Modeling for Traffic Analysis: State-of-Practice and
  Gap Analysis Report.
\newblock Technical Report FHWA-HRT-21-082, Federal Highway Administration.

\bibitem[{Zhou et~al.(2025{\natexlab{b}})Zhou, Kim, Ameli, Zhu, Honma, and
  Pendyala}]{zhou2025flowthrough}
Zhou, X.~S.; Kim, T.; Ameli, M.; Zhu, H.; Honma, Y.; and Pendyala, R.~M.
  2025{\natexlab{b}}.
\newblock Flow-through tensors: A unified computational graph architecture for
  multi-layer transportation network optimization.
\newblock \emph{Artificial Intelligence for Transportation}, 1: 100006.

\bibitem[{Zhu, Yang, and Chen(2021)}]{zhu2021vigor}
Zhu, S.; Yang, T.; and Chen, C. 2021.
\newblock {VIGOR}: Cross-View Image Geo-Localization Beyond One-to-One
  Retrieval.
\newblock In \emph{Proceedings of the IEEE/CVF Conference on Computer Vision
  and Pattern Recognition (CVPR)}, 3640--3649.

\end{thebibliography}

\section*{Appendix}
\appendix

\section{Network Understanding in Detail}
\label{app:understanding}
\begin{figure*}[t]
\centering
\includegraphics[width=\textwidth]{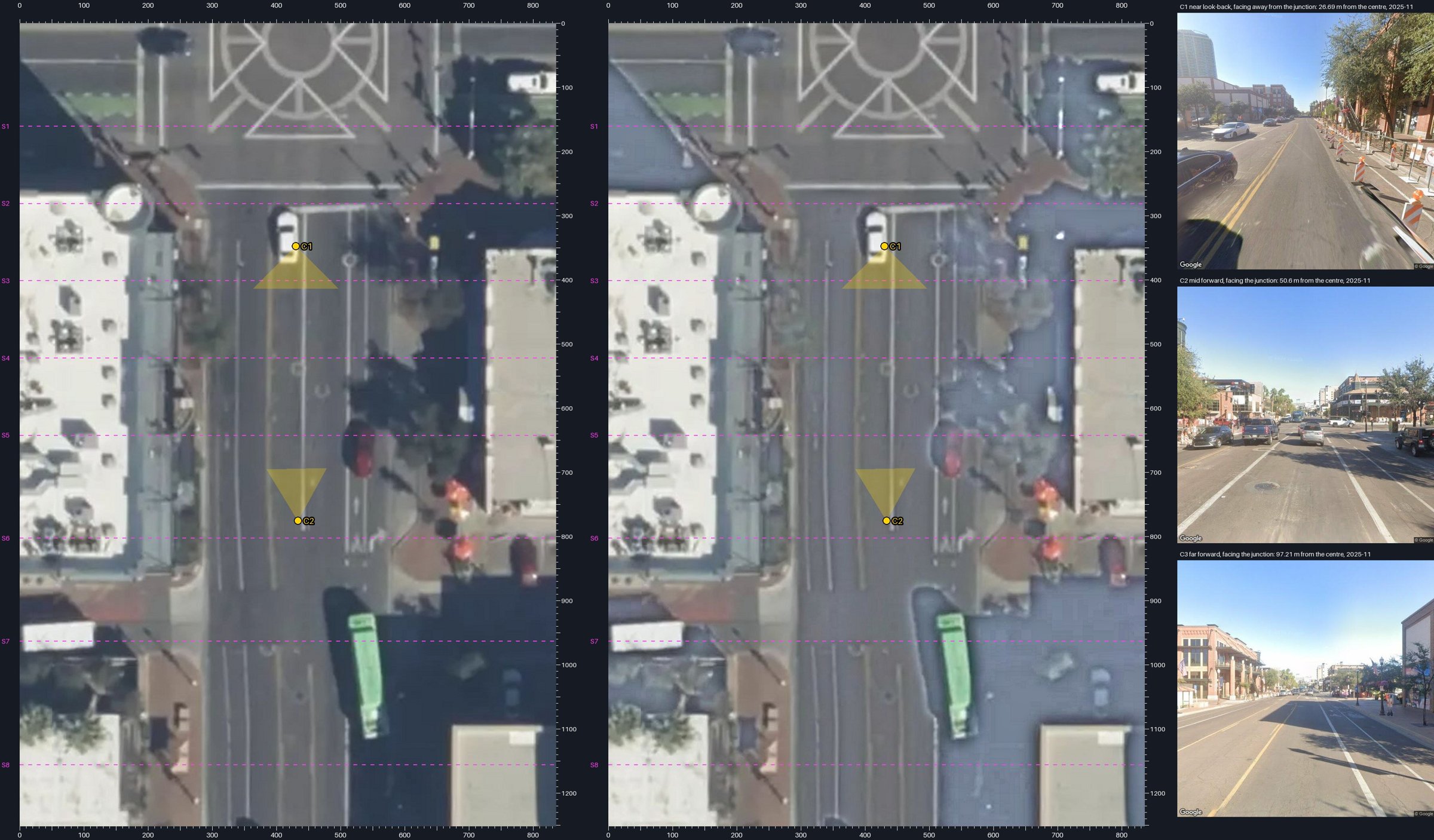}
\caption{Trace sheet of the western arm of the intersection in Fig.~\ref{fig:understanding}, as sent to the model. Left: the rotated crop with pixel rulers, stations S1--S8 (dashed, 10--72\,m from the junction centre) and cameras C1, C2 with their fields of view. Middle: the crop with shadows lightened. Right: the three street views, each captioned with its pose, distance and capture date. Satellite imagery: \textcopyright\,Mapbox, \textcopyright\,OpenStreetMap, \textcopyright\,Maxar; street-level imagery: \textcopyright\,Google.}
\label{fig:app-trace}
\end{figure*}

\begin{figure*}[t]
\centering
\includegraphics[width=\textwidth]{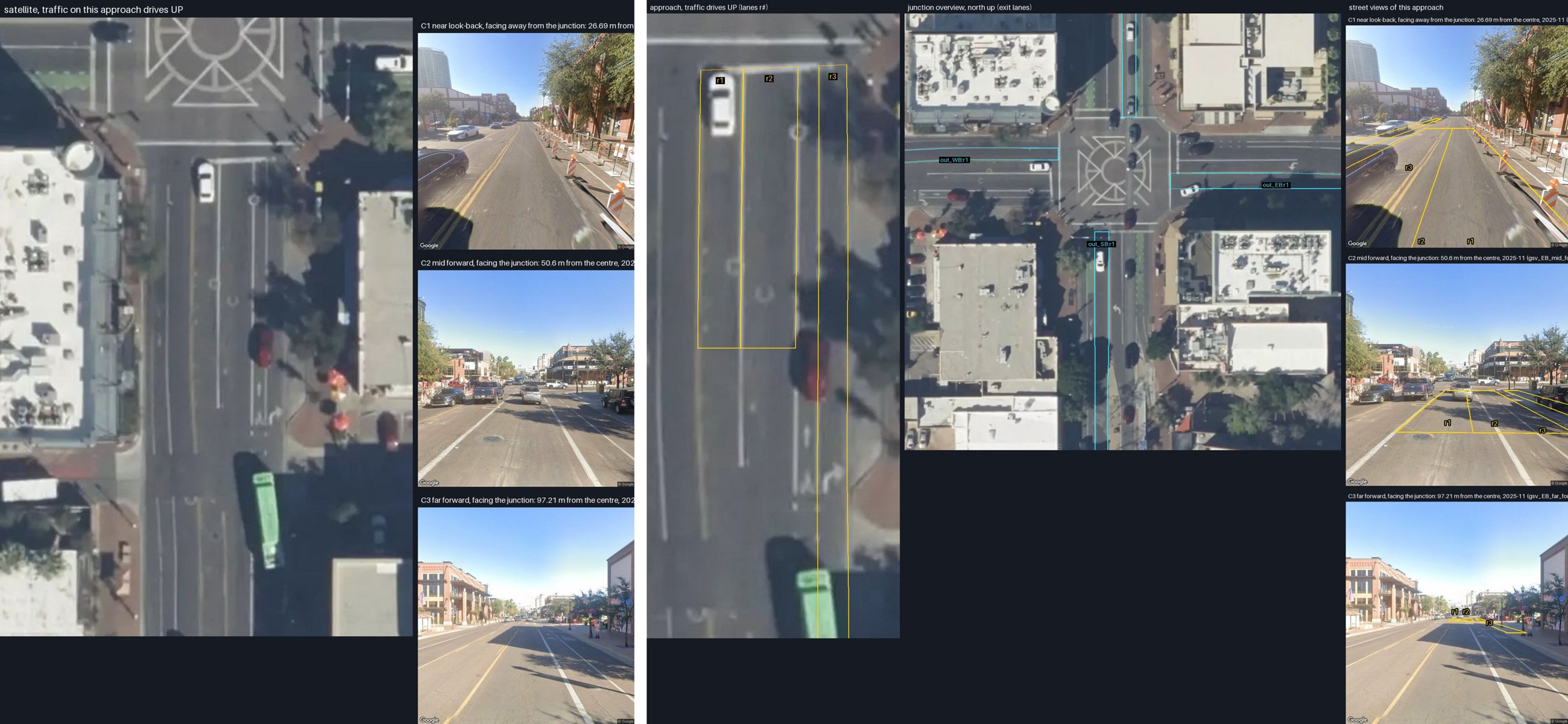}
\caption{Two more sheets of the same approach. Left: the blind reading, with the unmarked crop and the street views, and no trace or rulers. Right: lane use after the re-trace, with lanes r1--r3 outlined on the crop, the exit lanes on a north-up overview, and the lanes projected into the street views. Satellite imagery: \textcopyright\,Mapbox, \textcopyright\,OpenStreetMap, \textcopyright\,Maxar; street-level imagery: \textcopyright\,Google.}
\label{fig:app-sheets}
\end{figure*}

This appendix follows one intersection through the harness, request by request. Figures~\ref{fig:app-trace} and~\ref{fig:app-sheets} show the sheets of the approach in Fig.~\ref{fig:understanding} as the model received them, and Table~\ref{tab:requests} counts the requests of all six runs.

\paragraph{Arms and crops.} The legs of an intersection are its links open to motor traffic; legs whose bearings lie within 35$^\circ$ of each other form one arm. For each arm, code cuts a window from a zoom-19 satellite image, rendered at about 0.06\,m per pixel at our sites (512-pixel tiles at double density; the imagery's own ground resolution is coarser and not given by the provider), and rotates it so that inbound traffic drives up the image, with the junction centre at the top. The window reaches up to 78\,m from the junction centre, shortened in 2\,m steps where the image ends, and it extends 26\,m beyond the inbound and outbound links on either side. Eight stations lie at 10, 17.5, 25, 32.5, 40, 50, 60 and 72\,m from the junction centre and are drawn as dashed lines S1--S8; the rulers count crop pixels, with ticks every 10 and labels every 100. Where at least a fifth of the crop is in shadow, a second copy with shadows lightened is added and labelled as not true in colour. Code finds the shadow on the lightness channel with OpenCV: a pixel is shaded where its smoothed lightness is below 0.72 of the crop's upper-quartile lightness, and shaded pixels are brightened toward that level; a crop wholly in shadow is instead contrast-stretched by adaptive histogram equalisation.

\paragraph{Street views.} Each approach has three panoramas from the street-level archive, fetched once before any run: a near view about 25\,m from the junction centre (within 15--40\,m) looking back at approaching traffic, with a 90$^\circ$ field of view pitched 12$^\circ$ down, and mid and far views about 50\,m and 100\,m out facing the junction, with a 60$^\circ$ field of view; each is 640$\times$640 pixels. The heading follows the direction of travel on the link at the camera, reversed for the look-back view. To fetch a panorama on the approach's own carriageway of a two-way link, the query point is offset from the link by half the link's lane count in the network times 3.6\,m; this is the only use of a network lane count, and it places cameras without entering any prompt. Each camera inside the crop is drawn on the satellite crop as a dot, a label and a field-of-view wedge. Traced lanes are projected into the views with a pinhole camera at the recorded pose, a flat ground and a camera height of 2.5\,m.

\paragraph{Trace (one request per arm).} The prompt explains the frame (outbound carriageway on the left, inbound on the right, $y$ growing away from the junction) and the stations. The model returns the stop line as two points on the ruler frame (or, without a painted stop bar, the upstream edge of the crosswalk), the start of the exit, every longitudinal boundary of both carriageways as one $x$ per station with its kind (curb or pavement edge, lane line, centre line, median edge, bicycle line, buffer or parking edge, rail edge, unknown), and between each pair of boundaries the strip's surface (motor lane, shared rail and motor lane, shared bicycle and motor lane, bicycle lane, parking, median or buffer, shoulder, sidewalk or verge, rail or bus only, unknown) with a confidence and its evidence. Code checks that every value lies in the crop, that a stop line is given whenever the arm has an inbound carriageway, that boundaries are ordered and at least four pixels apart at every station, and that there is one strip fewer than boundaries. Code then builds lane regions: a boundary not drawn at the first station past the stop line merges the strips on either side; each region's width is measured at the first station past the stop line where both its boundaries are drawn; motor strips (the first three surfaces) become lanes r1, r2, \dots\ from the road centre, which is the driver's left. Outbound lanes are built in the same way past the start of the exit, and an exit's lane count is the number of its motor regions.

\paragraph{Blind reading (one request per approach).} The model sees the unmarked crop and the approach's three street views (Fig.~\ref{fig:app-sheets}, left), never the trace. It lists the lanes at the stop bar with their turns and evidence, any bicycle, rail or bus-only and parking lanes, a confidence for the count and whether the count is settled: ``If shadow, occlusion, image age or camera position prevents a reliable count, set count\_settled to false instead of guessing.'' A count is settled when the model says so and its confidence is not low.

\paragraph{Re-trace (at most once per arm).} Code lists the measured issues of the trace: a motor lane narrower than 2.5\,m or wider than 5\,m, a strip the trace itself marked low-confidence, and a bicycle strip at least 2.5\,m wide with a lane on both sides. The arm is re-traced if there is any issue or if a settled blind reading counts more inbound motor lanes than the trace; the reading is shown only in that case. The sheet draws the earlier boundaries in cyan as b1, b2,~\dots\ with the stop line and exit start, and projects the boundaries into the street views. The prompt asks the model to ``keep the trace where the pixels support it, even if it still differs from the separate reading or a strip stays unusually wide or narrow''. The new trace replaces the old one.

\paragraph{Lane use (one request per approach).} The sheet (Fig.~\ref{fig:app-sheets}, right) shows the approach crop, 45\,m long and enlarged twice, with the lanes outlined and labelled; a north-up overview of 110\,m with every exit lane outlined; and the street views with the lanes projected. The prompt gives each lane's width, the traced cross-section (for instance a bicycle lane of 1.7\,m between two lanes), each exit with its compass heading, signed turn angle and a plain description (straight ahead within 30$^\circ$; back the way the approach came beyond 150$^\circ$; otherwise left or right, sharp beyond 120$^\circ$), and the blind reading's count and arrows, to be used for arrows and signs, while a lane the reading does not count is judged from the images. It states the general rules of Section~\ref{sec:harness}, among them: ``The curbside lane may also turn right unless a permanent marking or fixed sign rules it out; this applies only to a strip that is a lane.'' For each lane the model answers whether it exists at the stop bar (yes, no, uncertain), its turns, whether they are observed or inferred, a confidence and evidence; for each turn, the exit, a status (candidate, unresolved, not proposed), the lane pairs from approach lane to exit lane and a reason. It may also list up to three lanes it sees at the stop bar that have no outline, report temporary traffic control (whether seen, which lanes, the effect and the evidence), and ask for detail crops. Code checks that a lane that does not exist has no turns, that exactly the lanes allowing a turn appear in that turn's lane pairs, that every allowed turn has a candidate or unresolved movement, and that only candidates have pairs. Temporary traffic control is recorded and changes nothing.

\begin{figure*}[t]
\centering
\includegraphics[width=\textwidth]{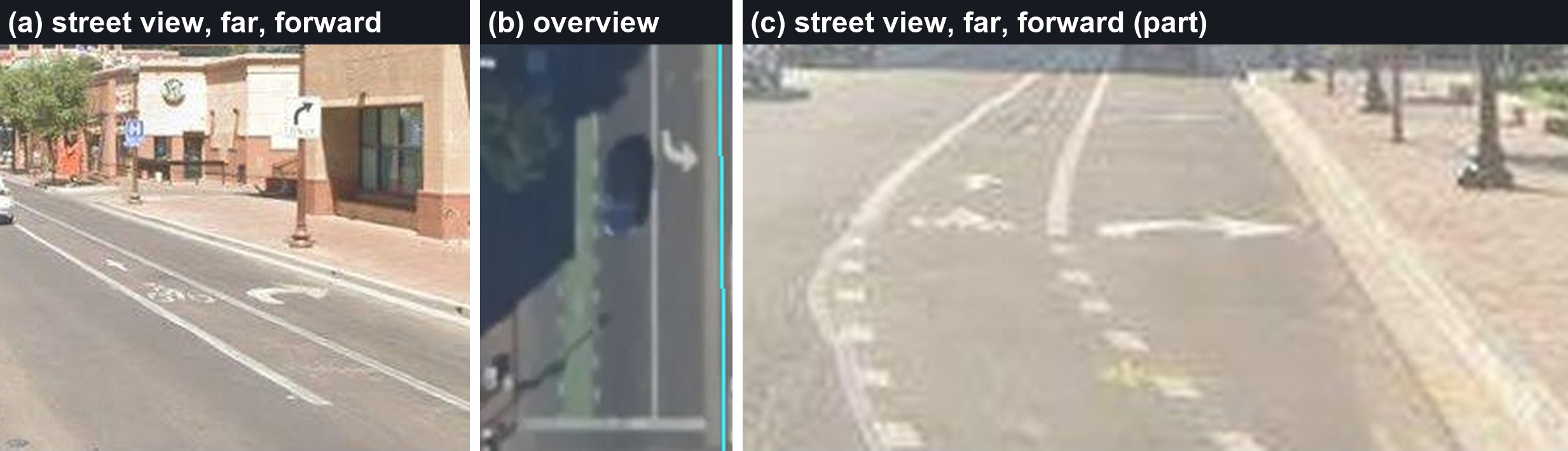}
\caption{Detail crops showing a sign or an arrow, as the model received them ((c) shows part of one). (a) GPT, westbound: a right-turn arrow beside the bicycle lane and a sign with a right arrow over ONLY; the answer keeps the curbside lane's right turn as observed. (b) Claude, southbound, north-up overview: a left-turn arrow in the median-side lane, kept as observed with high confidence. (c) Claude, westbound: a foreshortened right-turn arrow between the bicycle lane and the curb; the curbside lane's right turn, inferred before, is now observed with medium confidence. Street-level imagery: \textcopyright\,Google; satellite imagery: \textcopyright\,Mapbox, \textcopyright\,OpenStreetMap, \textcopyright\,Maxar.}
\label{fig:app-crops}
\end{figure*}

\paragraph{Detail crops.} ``If a small arrow, word or sign could settle a lane's use, ask for up to three native-resolution crops.'' Each crop names a source (the approach crop, the overview or one street view) and a box in fractions of it. Code cuts the box from the satellite image at source resolution, from the panorama at its native 640~pixels or from the rendered overview, enlarges it up to three times and sends the crops alone with the same question; the new answer replaces the old one and may ask for no further crops. Besides these, the model may crop any image it is sent, for instance to read ruler ticks or a thin line. Figure~\ref{fig:app-crops} shows three such crops.

\paragraph{Notes that appear only when their condition holds.} Each note is added to the lane-use prompt only for the approaches where code finds its condition, so every other request stays the same (Appendix~\ref{app:setup} gives their full text). A lane at least 2.5\,m wide beyond a bicycle strip: such a strip is normally where right-turning traffic waits, since curbside parking ends short of a junction, unless something at the stop bar keeps cars out. A bicycle strip at least 2.5\,m wide between two lanes: report a lane inside it if there is one. A lane given left and through by inference while another lane allows through: with several through lanes, the road-centre lane that takes left turns is usually a left-turn lane, which may lie outside the outlines on median-like paving; give only the turns that arrows or position show. A left exit with no left-turn lane: does the leftmost lane also allow left, or does a sign or marking ban it; give left where nothing rules it out. A lane kept against the answer: the lanes stay lanes because nothing independent confirms otherwise. A lane-wide lane about to be removed: keep it unless something at the stop bar keeps cars out.

\paragraph{Changing the set of lanes.} The lane-use answer cannot change the lanes itself. (i) If it lists lanes without an outline, the arm is traced again with these reports, and the new inbound lanes are adopted only if there are more of them; the exits stay as they were, and an adopted trace gets a new lane-use request. (ii) If it judges lanes not to exist, they are dropped only when the blind reading is settled and counts no more lanes than would remain and no rejected lane is 2.5\,m wide or wider; if a rejected lane is that wide, lane use is asked once more with the note above and that answer is final; otherwise lane use is asked again with every lane kept, each with a turn. (iii) If an exit lies to the left (a turn of 30--150$^\circ$ to the left) and no lane allows a left turn, the arm is re-traced to look for a left-turn lane on the road-centre side under rule (i); if that adds no lane, lane use is asked again with the left-exit note. (iv) If the first answer infers left and through for a lane while another lane allows through, lane use is asked again with the shared-lane note before (i)--(iii) apply.

\begin{figure*}[p]
\centering
\includegraphics[width=\textwidth]{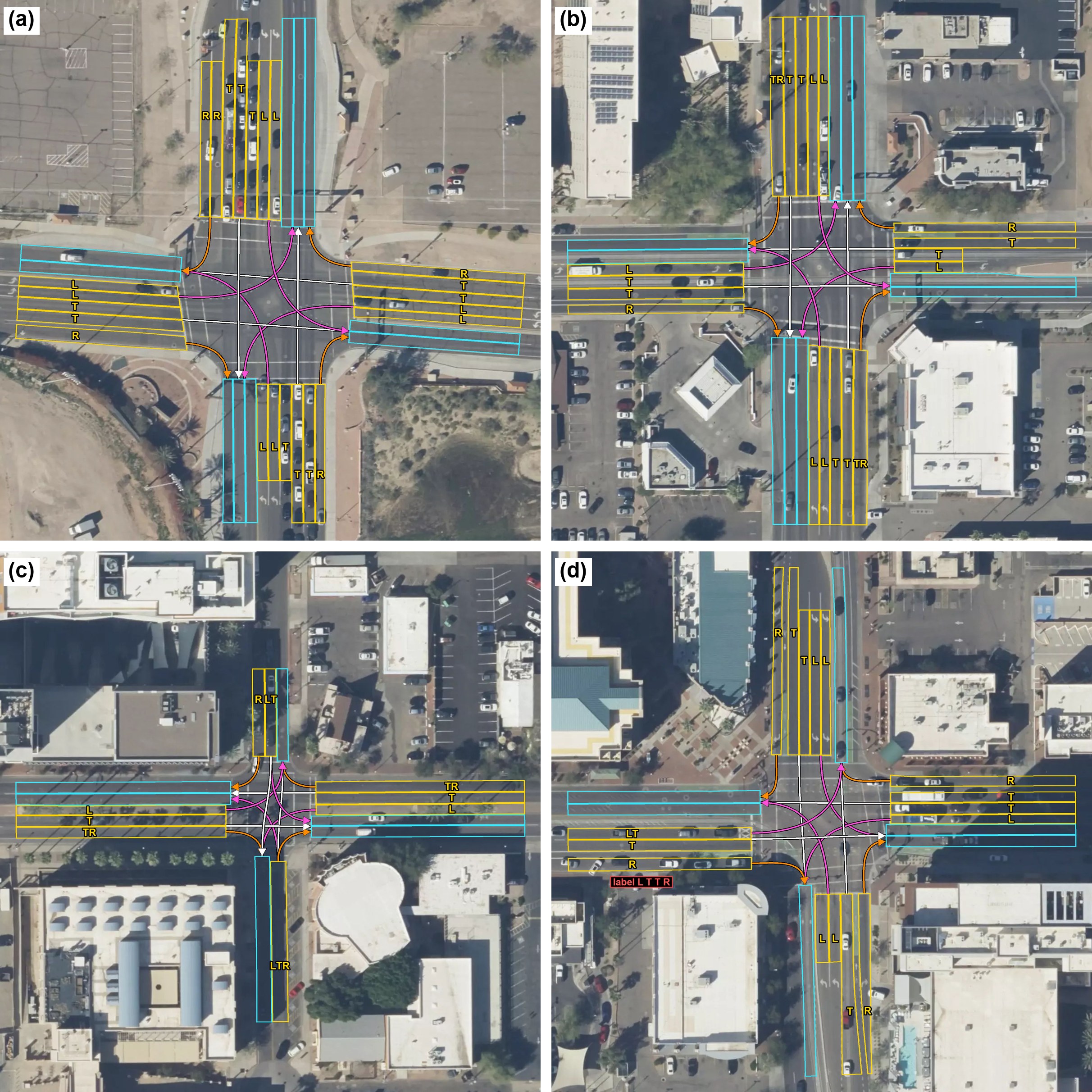}
\caption{Claude's first PAVE run at four of the 11 intersections, drawn from its lane and movement tables. Yellow: approach lanes at the stop bar with their turns, from the driver's left; cyan: exit lanes; arrows: movements from approach to exit lanes (left magenta, through white, right orange). Fifteen of the 16 layouts match the labels; in (d) the run reads the eastbound approach as three lanes, LT T R, where the label (red) has four. Satellite imagery: \textcopyright\,Mapbox, \textcopyright\,OpenStreetMap, \textcopyright\,Maxar.}
\label{fig:app-results}
\end{figure*}

\paragraph{Output.} The result is two tables keyed to the network. The lane table has one section per carriageway of each arm (inbound at the stop bar, outbound at the exit) with its link ID and, for each region, whether it is a supported lane, a lane rejected by lane use or not a motor strip, its lane index, width, outline and lane use. The movement table has, for each approach and turn, the status, the inbound and outbound sections and the lane pairs. These are the lane and movement records a lane-level network needs. Figure~\ref{fig:app-results} draws both tables for four intersections.

\paragraph{Repairs, budget, caching and replay.} An answer that fails its check is asked again with the error and the previous answer, at most twice; a request that still fails stops the run, which then writes no result. A run may make at most 40 requests per intersection. Each request goes to a fresh model instance that sees only that request; repairs and detail requests carry the earlier answer inside the prompt, not as conversation history. Every answer is stored under a signature of the image file, the prompt, the system prompt, the model and the request settings; an identical request is answered from the store, and any change to an image or a prompt makes a new request. A run can therefore be replayed exactly, or resumed after an interruption, and a run that may only use the store fails on the first missing answer. The code of each version is frozen by hash before its runs, and a run refuses to start if the hashes differ.

\begin{table}[t]
\centering
\small
\setlength{\tabcolsep}{3.5pt}
\begin{tabular}{@{}lcc@{}}
\toprule
Request & Claude & GPT \\
\midrule
Trace, one per arm & 44 & 44 \\
Blind reading, one per approach & 44 & 44 \\
Re-trace & 13/11/13 & 17/14/14 \\
Lane use, one per approach & 44 & 44 \\
Detail crops & 45/40/38 & 22/24/25 \\
Shared-lane re-ask & 3/1/2 & 4/3/3 \\
Trace for lanes without outline & 8/4/6 & 6/7/12 \\
Lane use with every lane kept & 1/0/0 & 0/0/0 \\
Lane-wide lane re-ask & 0/1/0 & 1/1/3 \\
Trace for a left-turn lane & 1/2/0 & 1/2/0 \\
Left-exit re-ask & 1/0/0 & 1/2/0 \\
Format repairs & 9/12/7 & 12/14/18 \\
\midrule
Total (11 intersections) & 213/203/198 & 196/199/207 \\
\bottomrule
\end{tabular}
\caption{Model requests of the six PAVE runs (first/second/third run), by step. The two trace rows also count the lane-use request that follows each trace. A single request makes one request per intersection, 11 in all. In the first run Claude's requests used about 0.12 million tokens each, including its reasoning and the crops it made, or about 2.4 million per intersection.}
\label{tab:requests}
\end{table}

\section{Scenario Editing in Detail}
\label{app:edit}

\paragraph{Answer and repair.} Each scenario asks the model once, with one rendered image and one prompt (Appendix~\ref{app:setup} prints them), and the answer is a single JSON object. The system prompt states that the model never draws, digitises or estimates coordinates: it answers only with IDs that exist in the tables it is given and with labels from closed catalogues. Anything the image and text do not establish, such as signal timing or demand, goes into a list of items not established (``unknown is not zero''). If a gate rejects the answer, one repair request repeats the query and the previous answer, lists one line per defect (gate, where, message) and asks for a corrected object with the same keys. In our runs the gates rejected only GPT's three first answers for S3, whose westbound sections contained a junction (P6), and each repair passed them. Built without the gates, these sections pass the check on side roads inside a section, which looks only at the cells (the lane segments of the S3 network); two of the three leave trucks out of the general lane, before and after the repair, and fail C5, since the gates check cars and slow vehicles but not trucks. The facts an answer expects to hold afterwards are recorded but never decide a check. Table~\ref{tab:scenarios} lists what each scenario shows and asks.

\begin{table*}[t]
\centering
\small
\begin{tabular}{p{0.09\textwidth}p{0.26\textwidth}p{0.26\textwidth}p{0.26\textwidth}}
\toprule
 & S1: add a lane (US-60) & S2: diamond to DDI (SR~347 / I-10) & S3: passing lanes (US-60) \\
\midrule
Image & the whole subarea, and $\pm$1.5\,km around the freeway crossing with each road number shown once; 6$\times$6 grid
 & satellite panel and a drawing of the network within $\pm$230\,m of the ramp terminals, every link and node ID labelled; 6$\times$6 grid
 & the briefing slide of the scenario, and the corridor with a station every 100\,m on each carriageway and junctions marked \\
Tables & links: ID, from and to node, link type, lanes, corridor, length
 & nodes: ID, type, $x$, $y$; links: ID, from and to node, facility type. Lanes, speeds, capacities and road names withheld
 & corridor length, stations, junctions; one lane in each direction today \\
Answer & the links to widen with their lanes before and after, and the periods
 & facility class, arterial axis, two terminals with six bound roles each, the operation
 & one section per direction (west and east station) and the use of the passing and general lanes \\
Gates & P0--P9 (10) & P0--P8 (9) & P0--P8 (9) \\
Checks & C1--C6 (6) & C1--C6 and DDI-1--3 (9) & R1--R3 and C1--C7 (10) \\
\bottomrule
\end{tabular}
\caption{What each scenario shows the model, what it asks and how the answer and the edit are checked. Each scenario type has its own gates and checks; their names restart at P0 and C1.}
\label{tab:scenarios}
\end{table*}

\paragraph{Interchange gates (S2).} A \emph{port} is the node where a role's link meets its ramp terminal, and a \emph{connector} joins two ports. P0: the answer parses as JSON. P1: required keys and types are present, the class, operation and facts come from the catalogues, each terminal binds exactly the six roles, each with a link ID and its from and to nodes, and terminal names are distinct. P2: every bound link and every axis node exists. P3: the from and to nodes copied for each link match the table. P4: each role's link has the right facility type, the six port nodes are distinct, the six diamond connectors exist between the ports and no other connector joins them. P5: the bridge is continuous (the bridge link leaving one terminal is the one entering the other, or an unbranched chain of links). P6: the grid cell the answer names for a terminal is the cell nearest to its ports. P7: the class agrees with the class read from the tables (any crossover node means a DDI; one off- and one on-ramp at each of two terminals a diamond; two of each at one terminal a SPUI) and the terminal count fits it. P8: the operation is implemented, starts from the answer's class, has a template, and names bound terminals. A later gate runs only when the earlier ones pass.

\paragraph{Lane-addition gates (S1).} P0 parse; P1 schema (the operation adds a positive number of lanes; the periods are AM, midday and PM); P2 every link exists; P3 from and to nodes match; P4 the lanes before equal the table and the lanes after add the lanes added; P5 the opposite direction of every link is named too; P6 the links form one chain in each direction; P7 the road is the one the brief names and the link type is a road type (not a connector); P8 the grid cell holds the links' mean midpoint or is next to it; P9 each direction is 0.3--0.8 mile long.

\paragraph{Passing-lane gates (S3).} P0 parse; P1 schema (one eastbound and one westbound section); P2 each station exists on its carriageway; P3 the west station lies west of the east station; P4 each section is 1.2--2.0\,km long; P5 the section has one lane today; P6 no junction lies inside a section, more than 50\,m from its ends; P7 the eastbound section ends no more than 0.3\,km east and no more than 1.0\,km west of where the westbound one starts; P8 the passing lane carries cars but not slow vehicles, and the general lane carries both.

\paragraph{Where the thresholds come from.} The length gates turn the brief's own figures into ranges: about half a mile for S1 (P9), one mile for each passing lane (P4, 1.61\,km) and mileposts at which the two passing lanes meet (P7). The other distances (150\,m for same-leg returns in C2; in S3, 50\,m in P6, 3\,m in C4 and 80\,m in C7) are our settings. None is taken from the agency's build networks.

\paragraph{Interchange checks (S2).} C1 table integrity: both ends of every link exist, lengths are positive, link geometry starts and ends at its nodes, and each movement's inbound link ends and outbound link starts at its node. C2 external reachability: between boundary nodes of the approach roads (pairs closer than 150\,m dropped as same-leg returns), no origin--destination pair that was reachable before the edit becomes unreachable; in S2, 14 of 17 pairs are reachable before and after. C3 movements: within each terminal's control group (the links its signal controls), every required port-to-port path exists and no prohibited one does. C4 grade separation: no arterial link or connector touches a freeway node, and no two links on the same layer cross without a shared node. C5 the freeway mainline is unchanged. C6 nothing outside the edited terminals changes. The templates add their own checks: for the DDI, each through path passes exactly one crossover (DDI-1), traffic runs on the left between the crossovers (DDI-2) and the two crossover nodes stay separate (DDI-3); for the SPUI, one signal group (SPUI-1) and one link per movement (SPUI-2); for roundabout terminals, counter-clockwise circulation (RBT-1), a one-way ring (RBT-2) and yield-controlled entries without signals (RBT-3).

\paragraph{Lane-addition and passing-lane checks.} S1: C1 table integrity, C2 topology unchanged, C3 only lanes and lane-based capacities change, C4 link capacity scales with the lane count, C5 every zone pair (an origin and a destination of the demand) reachable before is still reachable (2{,}674 of 3{,}422 pairs, before and after), C6 no lane-change columns of the planning scenario are left in the table. S3: R1 each section lies on one carriageway in one direction, R2 it has one lane today, R3 no junction inside it; C1 integrity, C2 nothing existing moved or rewired, C3 lane continuity with the merge at the section end, C4 new nodes more than 3\,m from existing ones, C5 the two zones still reach each other for each vehicle class, C6 lane use as asked, C7 no new U-turn within 80\,m of a section end.

\paragraph{Templates.} The three interchange templates share six roles at each ramp terminal: the arterial link arriving from the outer leg (approach in), the arterial link leaving toward it (departure out), the arterial links arriving from and leaving toward the other terminal across the freeway (bridge in and out), the off-ramp and the on-ramp. A diamond terminal requires the two through movements, the two right turns (approach to on-ramp, off-ramp to departure) and the two left turns (bridge to on-ramp, off-ramp to bridge), and prohibits returns along the arterial and off-ramp to on-ramp. In a diverging diamond the two directions of the crossing road swap sides at a \emph{crossover} before each terminal, so that traffic runs on the left between the crossovers and turns left onto the on-ramps without crossing oncoming traffic. The DDI template swaps the positions of the two bridge ports at each terminal, splits the two through connectors where they cross into two distinct crossover nodes with signals, and marks the turn connectors for control to be verified; it stops with an error unless the through connectors are straight and cross. The SPUI template merges the two terminals into one signalised point midway between them, moves the arterial and ramp ports outward, and adds ten connectors (two through, four right, four left) in one signal group. The roundabout template replaces each terminal's connectors with a one-way ring of 30\,m radius with yield-controlled entries, and requires every entry--exit path.

\paragraph{Design guidance.} After the checks, each design is rated against design guidance, item by item, as met, not met, schematic, measured, needing data, or for information. The items come from the Arizona DOT Roadway Design Guidelines \citep{adot2021rdg} first and then national guides: the DDI guide \citep{cunningham2021ddi}, the SPUI report \citep{messer1991spui} and the roundabout guide \citep{nchrp2023roundabouts}, among others. In S2 the crossover angle, which the guide wants at 45$^\circ$ or more, measures 75.3$^\circ$ and 81.2$^\circ$ at the two terminals.

\paragraph{Outputs and comparison with the agency's design.} An interchange edit writes an after-network (node, link and movement tables in GMNS) and a change table that lists every node, link and movement as unchanged, modified, added or deleted; S1 and S3 write their edits into the agency network's own link table. S1: the lanes per period of the named links against the no-build lanes plus the build's lane changes; with PAVE both directions of the named stretch go from two to three lanes in all three periods, as in the agency's design. PAVE also scales the capacity of the fourth period, which the brief does not name and the agency's design leaves at two lanes. S2: the distance from each crossover to the agency's crossing point near its terminal, and 16 movements (eight at each terminal: the six diamond movements, the ramp-to-ramp movement and the U-turn) traced in the agency's DDI network. The crossover position is the main measure, since the unchanged diamond agrees with 15 of the 16 movements as well; with PAVE 15 agree, the exception being the ramp-to-ramp movement that the agency's design drops. S3: the overlap in kilometres along the corridor between each passing lane and the agency's. With PAVE the eastbound lane runs km 5.10--6.70 against 3.93--5.88 in the agency's design (0.78\,km), and the westbound lane km 6.80--8.40 against 6.80--8.38 (1.58\,km). The brief asks for one mile each way from a milepost that the briefing slide places at a side-road junction; the agency's westbound lane starts there, but its eastbound lane ends 0.9\,km short of it and is 1.2 miles long. GPT's first answers, which the gates rejected, ran the westbound lane over km 6.70--8.30 (1.50\,km).

\paragraph{Control conditions.} Without the harness (b), each run is a fresh instance of the model working as a general agent in a folder of its own. The folder holds the brief (the scenario, image and network facts of PAVE's request, without PAVE's answer format), the same image, the complete No-Build network files and a page of file conventions (units, field meanings, how lane use is coded). The agent may read every file and write and run Python, and must write the complete Build network in the input format with a note on what it changed. It is told to keep existing IDs and records unless the edit requires changing them, to change nothing outside the project and to keep every No-Build trip possible; it is not shown the gates, templates, checks or the agency's design. It sees more than PAVE's model, which receives only the columns its question needs. Each run was limited to its folder. An audit of Claude's tool calls found two S3 runs that wrote a summary of their own No-Build network to a temporary file outside the folder; no run opened anything else. Before scoring, differences of format are removed: columns are matched by name and numbers by value, and S2 links without a drawn line, length or layer get the straight line, its length and the No-Build layer of their facility type. The pipeline's S3 checks rely on the stations PAVE's answer names, so a general version finds the edit from the tables, assigns each new lane to the carriageway its new links attach to, and runs R1--R3 and C1--C7 with the same thresholds; its rule for that assignment was corrected twice while scoring the agents, and on the networks PAVE built it returns the same verdicts as the pipeline, while an unedited network fails it. Each correction removed false failures and none added one: under the first version the agents passed 7, 7 and 6 of the ten S3 checks (Claude) and 5 of ten in each run (GPT), against 8, 8, 9 and 8 under the final version, and PAVE's verdicts are the same under all three versions. The S2 conventions page gave link length in metres where the files use kilometres; the agents that noticed kept kilometres, and no check depends on it. Without gates (c), the first answer of each PAVE run is built as it is, with every gate turned into a record of its defects. The checks and the comparison use reference terminals read from the link names, never from an answer, so a wrong binding shows up as failed checks.

\paragraph{Paths and movement records.} The paths of C2, C3 and the 16 movements follow the movement records, which the conventions page defines as the allowed movements at a node: a path passes a node that has records only along a recorded movement. The scorer frozen before these runs searched the links alone. On the agents' shared crossover nodes, whose records allow only the two through movements, it found U-turns that the records forbid and failed C3 in five agent runs (Claude's second and third, all three of GPT's); with the records, these runs pass C3, and their movements agree with the agency's on 15 of 16 instead of 13. No PAVE run and no other check changes. In the agency's lane-level DDI the two directions cross without a shared node.

\begin{figure*}[p]
\centering
\includegraphics[width=\textwidth]{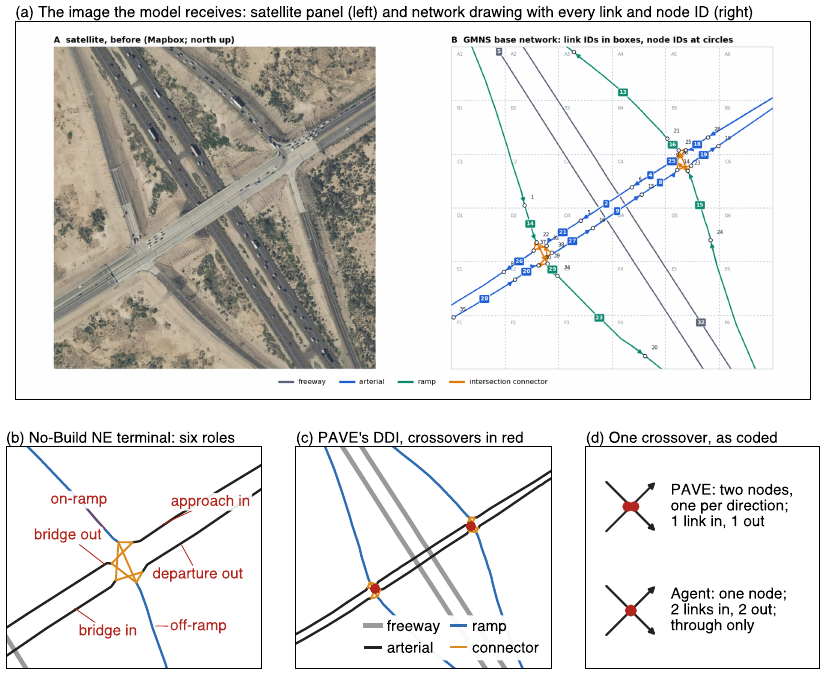}
\caption{Scenario S2. (a) The image the model receives: the satellite panel and the network drawing with every link and node ID on a 6$\times$6 grid. (b) The six roles bound at the NE ramp terminal of the No-Build diamond (local metres). (c) The DDI that PAVE's template builds from the twelve roles; red dots are crossovers. (d) PAVE codes a crossover as two nodes, one per direction; every agent run codes it as one shared node that allows only the two through movements, so it passes the movement checks but fails DDI-3. Satellite imagery: \textcopyright\,Mapbox, \textcopyright\,OpenStreetMap, \textcopyright\,Maxar.}
\label{fig:edit}
\end{figure*}

\paragraph{Binding from the tables alone.} The S2 request shows node IDs, types and coordinates and link IDs, end nodes and facility types. A short script written for a diamond interchange binds the twelve roles from these columns alone: a terminal is a group of port nodes joined by intersection connectors, its off- and on-ramp are the ramp links that end and start there, and of its two arterial links in each direction the bridge link is the one whose chain of arterial links reaches the other terminal. It agrees with the reference on all twelve roles (Fig.~\ref{fig:edit}b). The model needs no such rule for each interchange form, but we have not tested tables that are incomplete or wrong, where the image would have to decide.

\paragraph{Failed checks of the agents.} Table~\ref{tab:agentfail} sorts the checks the agents fail into design or coding errors and conventions of the checks. In S1 all six agent runs leave the fourth period (a time slice of the model day) unscaled, as does the agency's design, which changes the lanes only in the three periods the brief names; one run also changes the load factors (parameters of the links' travel-time function). In S2 every agent run codes each crossover as one node shared by both directions (Fig.~\ref{fig:edit}d), whose movement records let no traffic turn there, so the template's check that crossover nodes stay separate (DDI-3) fails. Like PAVE, the agents keep the ramp-to-ramp movement that the agency's design drops, and five of the six runs agree with the agency on 15 of the 16 movements; one Claude run leaves its new connectors outside the terminals' signal groups, so its through movements fail C3. In S3 two of Claude's runs leave each new lane dead-ending at both ends, within 2.2\,m of a junction node; GPT's lanes all run past the junction before merging back; and the four runs that add the passing lane on the inside renumber the lanes of about 460 existing nodes, as the network's lane numbering requires, which the check on existing records counts as a change.

\begin{table}[t]
\centering
\small
\setlength{\tabcolsep}{3pt}
\begin{tabular}{@{}lp{0.5\columnwidth}ccl@{}}
\toprule
 & Failed check & Claude & GPT & Kind \\
\midrule
S1 & C4: fourth period's capacity not scaled & 3 & 3 & convention \\
 & C3: load factors changed & 1 & 0 & error \\
S2 & DDI-3: one crossover node shared by both directions & 3 & 3 & convention \\
 & C3, C6, DDI-1: connectors outside the signal groups & 1 & 0 & error \\
S3 & C3, C4: new lanes dead-end; a new node 2.2\,m from a junction node & 2 & 0 & error \\
 & R3: passing lanes run past a junction & 0 & 3 & error \\
 & C2: existing lanes renumbered by an inside lane & 1 & 3 & convention \\
\midrule
S3 & PAVE, C5: trucks left out of the general lane & 0 & 2 & error \\
\bottomrule
\end{tabular}
\caption{Failed checks of the agent runs (no harness): runs out of three in which each fails, and PAVE's two failures for comparison. Error: a design or coding error. Convention: the check's own convention, which the agency's design or the network's numbering does not follow.}
\label{tab:agentfail}
\end{table}

\section{Labels and Scoring}
\label{app:scoring}
\paragraph{Labels.} Every approach was labelled by hand from the satellite image and the street-level panoramas: the motor lanes that reach the stop bar, in order from the driver's left, the turns each lane allows (left, through, right) and the number of lanes of each exit. U-turns are not labelled or scored. The 11 intersections have 44 approaches and 43 exits. The approaches have one to seven lanes at the stop bar (15 have at most two, 13 three, 7 four and 9 five or more), and 25 have a lane shared by two turns.

\paragraph{Metrics.} An approach is scored on four items. \emph{Count}: the number of motor lanes at the stop bar. \emph{Layout}: the list of per-lane turn sets, from the driver's left, equals the label (the main metric). \emph{Turn set}: the union of the turns over all lanes equals the label's. \emph{Turn lanes}: for each of left, through and right, the number of lanes that allow it equals the label's. For PAVE we also score the lane count of each exit. An approach a run left without output counts as wrong. Every condition is averaged over its three runs or repeats.

\paragraph{Statistics.} A difference between PAVE and a single-request condition is the mean over approaches of PAVE's mean score over its three runs minus the mean score of the three repeats. The 95\% interval runs from the 2.5th to the 97.5th percentile of 10{,}000 bootstrap draws of the 11 intersections with a fixed seed. The $p$ value is a two-sided exact sign-flip permutation test over all $2^{11}$ sign patterns of the per-intersection sums of these differences. Each comparison is reported on its own, without a correction for multiple comparisons. For the layout, PAVE against the satellite and the street-view single request gives $p=0.014$ and $p=0.009$ with Claude and $p=0.006$ and $p=0.001$ with GPT, and PAVE against its blind reading $p=0.004$ with either model.

\paragraph{OpenStreetMap.} For approaches whose inbound link has OpenStreetMap turn-lane rows (29 of 44), the \texttt{turn:lanes} value of the way is read cell by cell from the driver's left, keeping shared cells such as left;through, with a blank cell read as through, and scored in the same way. A second variant also lets the rightmost lane turn right when its cell carries no marking and the junction has an exit to the right.

\paragraph{Blind reading and rule table.} The blind reading lists each lane with the turns its arrows show, or none. Read as reported it gets 4.3 (Claude) and 4.0 (GPT) layouts right, means of three runs; with a lane without arrows read as through, 12.7 and 12.0; with the rightmost such lane also turning right where the approach has an exit to the right, 21.3 with either model, the values in Table~\ref{tab:understanding}. The rule table takes an approach's lane count and the turns its exits allow, from the network geometry: the leftmost lane turns left, the rightmost goes through and right, every other lane goes through, and an approach without a through exit splits its lanes between left and right. With the labelled counts it gets 19 layouts right, with PAVE's counts 15--18 (Claude) and 17--18 (GPT), and with an exclusive right-turn lane on approaches of three or more lanes 17 or fewer. The other differences in Section~\ref{sec:experiments}, between the two single-request variants and between PAVE and its blind reading, use the same bootstrap and permutation test.

\paragraph{Errors and the steps that catch them.} Table~\ref{tab:failures} sorts the wrong layouts of every condition by kind, against the label: fewer or more lanes; the same lanes read mirrored, as if from the other side; and other wrong turns on a correct count. For PAVE it then follows each lane count the trace gets wrong: whether the blind reading (a settled reading that counts more lanes) or a measured issue (a width out of range, a low-confidence strip, a wide bicycle strip between lanes) flags it for the re-trace, and which step fixes it. The sorting is done by code from the scored rows and each run's stage record. The lower rows add up: with Claude, the trace leaves 7.3 counts per run wrong; the re-trace and the lane-set rules fix 2.0 each, the re-trace breaks 2.3 others, the lane-set rules fix 1.7 of these later, and 4.0 remain wrong (each mean is rounded on its own).

\paragraph{Why the re-trace and the detail crops stay.} The re-trace is a guard, and the mean hides the run it guards. A run cannot tell that its trace is poor, and the network gives no sign of a wrong count. In GPT's first run the trace got 68.2\% of the counts right, and the re-trace fixed nine counts and broke one. It costs 11--17 requests per run, about 7\% of all requests, and the lane-set rules later fix 9 of the 15 counts it breaks in the six runs. The detail crops make the evidence firmer: in 16 rounds a turn is now read from an arrow or held with more confidence (Fig.~\ref{fig:app-crops}c). PAVE's output is a proposal with an evidence record for a modeller to review, so firmer evidence counts even when the turn stays the same.

\begin{table*}[t]
\centering
\small
\setlength{\tabcolsep}{6pt}
\begin{tabular}{@{}lcccccc@{}}
\toprule
 & \multicolumn{3}{c}{Claude Opus 5.5} & \multicolumn{3}{c}{GPT-6.1 Sol} \\
\cmidrule(lr){2-4}\cmidrule(lr){5-7}
Wrong layouts & Satellite & Sat.\ + street & PAVE & Satellite & Sat.\ + street & PAVE \\
\midrule
Fewer lanes & 11.7 & 9.7 & \textbf{1.7} & 13.3 & 18.0 & \textbf{4.7} \\
More lanes & 4.0 & 4.7 & \textbf{2.3} & 1.0 & 0.3 & \textbf{0.0} \\
Mirrored & 1.0 & 0.3 & \textbf{0.0} & 0.7 & 0.7 & \textbf{0.0} \\
Other turns & 3.3 & 7.3 & \textbf{0.0} & \textbf{1.0} & 3.3 & 1.7 \\
\midrule
\multicolumn{7}{@{}l}{\emph{PAVE, step by step}} \\
Count wrong after the trace & & & 7.3 & & & 9.7 \\
\quad flagged by the blind reading & & & 1.3 & & & 3.0 \\
\quad flagged by a measured issue only & & & 3.7 & & & 4.0 \\
\quad not flagged & & & 2.3 & & & 2.7 \\
\quad fixed by the re-trace & & & 2.0 & & & 5.0 \\
\quad fixed by the lane-set rules & & & 2.0 & & & 1.7 \\
Counts the re-trace broke & & & 2.3 & & & 2.7 \\
\quad fixed later by the lane-set rules & & & 1.7 & & & 1.3 \\
Counts the lane-set rules broke & & & 0.0 & & & 0.3 \\
Count wrong at the end & & & 4.0 & & & 4.7 \\
Layout wrong at first lane use & & & 7.3 & & & 9.7 \\
\quad fixed by detail crops & & & 0.0 & & & 0.3 \\
\quad fixed by lane-set rules, re-asks & & & 3.3 & & & 3.0 \\
Layout wrong at the end & & & 4.0 & & & 6.3 \\
\bottomrule
\end{tabular}
\caption{Wrong layouts by kind and the steps that catch them. Every number counts approaches (of 44), averaged over three repeats (single requests) or three runs (PAVE). Upper rows: approaches whose layout is wrong, by how it differs from the label. Lower rows, PAVE only: approaches whose lane count is wrong after the trace, how many of these each check flags and each later step fixes, counts that a step makes wrong, and the counts still wrong at the end; then the same for layouts from the first lane-use answer on. A lane count flagged both by the blind reading and by a measured issue is counted under the blind reading. Bold: the fewest in each upper row for each model.}
\label{tab:failures}
\end{table*}

\section{Lane Tables as Step 0 of a Transportation Model}
\label{app:model}
A transportation model generates trips, distributes them between zones and assigns them to routes, or simulates vehicles along them, on a road network that it takes as given. Building that network comes before these steps, as a step 0 (Fig.~\ref{fig:stepzero}), and PAVE's output can be written in the model's terms. For an approach $a$ with lanes $k=1,\dots,n_a$ from the driver's left and an exit $b$, let $u_{akb}=1$ when lane $k$ may serve the movement from $a$ to $b$, and $u_{akb}=0$ otherwise; the lane pairs add which exit lane each such movement enters. \citet{wu2018hierarchical} write trip generation, OD flows, route flows and link flows as the layers of one computational graph. The lane table adds a layer beneath the link flows, in which the flow $X_{akb}$ of movement $(a,b)$ in lane $k$ satisfies \[ \textstyle\sum_k X_{akb}=X_{ab}, \qquad 0\le X_{akb}\le u_{akb}\,c_{ak}, \] with $X_{ab}$ the flow of the movement and $c_{ak}$ the capacity of the lane. Flow passes only where $u_{akb}=1$: the lane table is the support of the flow-through tensor $X_{akb}$ \citep{zhou2025flowthrough}. A turn that no lane serves removes every route that uses it, and a lane too many or too few changes the capacity of the approach, hence its travel time and the routes chosen. In a simulation, the same table gives the lanes each turn may use and the lane-to-lane connections. Network understanding estimates this supply for the network as built; a scenario edit changes it to a design, on which the same graph would run with forecast demand.

Run backward, the graph returns the gaps between modelled and observed flows, such as turning counts, to the layers above as gradients \citep{wu2018hierarchical}: to demand, route choice and capacity. They do not reach $u$, which is discrete; changing one entry adds or removes routes. A gap can still send an approach back to be read again. A movement with counted flow that no lane serves, $\hat X_{ab}>0$ while $\sum_k u_{akb}=0$, rejects the table outright. A lane too many or too few, or a turn moved from one lane to its neighbour, changes only capacity, so it shows in counts only where the approach runs near capacity, and elsewhere only in travel times. This is a check from outside the images, which PAVE's remaining errors need: its wrong layouts are legal ones (below). We loaded no demand on PAVE's tables, so no such check is run here.

\begin{figure}[!t]
\centering
\includegraphics[width=\columnwidth]{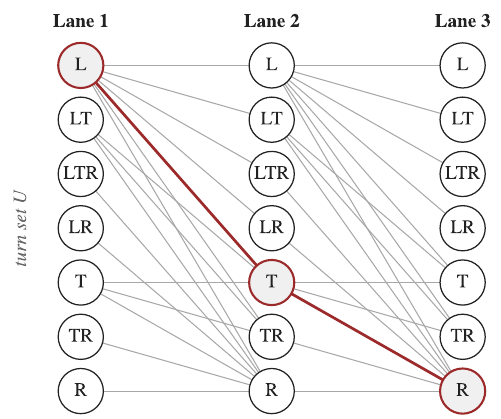}
\caption{Lane-state network of a three-lane approach: columns are lanes from the driver's left, nodes the seven turn sets. Grey arcs: the 17 pairs the no-crossing rule allows in adjacent lanes. Red: the labelled layout L, T, R of the approach in Fig.~\ref{fig:understanding}. Decoding would find the cheapest path under node costs $\ell_k(U)$ and arc costs $\ell(U',U)$; none is run here.}
\label{fig:lanestate}
\end{figure}

Within one approach, the layout itself can be written as a path. Each lane takes one of the seven non-empty sets of turns from $\{\mathrm{L},\mathrm{T},\mathrm{R}\}$, and movements from adjacent lanes must not cross \citep{wong2003lane}: a set $U$ may follow $U'$ to its right only if no turn in $U$ lies to the left of a turn in $U'$, which holds for 17 of the 49 pairs (Fig.~\ref{fig:lanestate}). The rule leaves 31 of the 343 layouts of a three-lane approach and 127 of the 823{,}543 of a seven-lane one. With ${\ell_k(U)=-\log\Pr(o_k\mid U)}$ the cost of reading the evidence $o_k$ of lane $k$ as $U$, and ${\ell(U',U)=-\log\Pr(U\mid U')}$ the cost of $U$ beside $U'$, the most likely layout follows from a dynamic-programming recursion over the allowed pairs, \[ C_k(U)=\min_{U'}\bigl[C_{k-1}(U')+\ell(U',U)\bigr]+\ell_k(U). \] In our runs the rule changes nothing: none of the 264 final layouts (44 approaches, three runs each of two models) breaks it. What decoding could add is a probability for each approach and its next most likely layouts, telling a modeller which approaches to check. That needs readings calibrated as probabilities, which PAVE's marks of doubt are not (Section~\ref{sec:limits}). We did not decode.

\raggedbottom  
\section{Prompts}
\label{app:setup}
\paragraph{Network understanding prompts.} Every PAVE request for network understanding sends the system prompt below, a fixed instruction that precedes the request, with one image and one prompt. The prompts are printed here as the code builds them. Text in italics stands for values that change from request to request, such as the intersection, the link and lane IDs, the lane widths and the photo captions; Appendix~\ref{app:understanding} describes the images.

\begin{promptbox}{System prompt (network understanding)}
You are a transportation-image analyst. Distinguish visible evidence, geometric inference and unknowns.\par
Region IDs and proposed boundaries are annotation aids, not semantic labels or ground truth.\par
Return structured JSON only. Never invent visible arrows, image dates or confirmed legal permissions.
\end{promptbox}

The trace prompt for an arm with both carriageways, shaded parts and street views:

\begin{promptbox}{Trace (one request per arm)}
Trace the \emph{arm} arm of the signalised intersection \emph{intersection}.\par
The image is a satellite crop rotated so that traffic ENTERING the junction on this arm moves UP. The crop starts at the junction centre, so the junction, its crosswalk and the stop bar lie near the TOP edge. Traffic keeps to the right.\par
It holds the inbound carriageway (\emph{inbound link}: traffic entering the junction, moving UP, on the right of the road centre) and the outbound carriageway (\emph{outbound link}: traffic leaving the junction, moving DOWN, on the left of the road centre). A second panel shows the same crop with its shaded parts lightened (colours there are not true; dark cars and new asphalt are lightened too) with the same rulers; use it to see paint in shadow. On the right are street views of the inbound approach (\emph{C1: pose, distance from the junction centre, capture date}; ...); their cameras are marked in yellow on the crop with their field of view (positions from street-view metadata, off by a few metres at most). Use them to see lane lines, arrows, bicycle lanes and curbs that shadow or vehicles hide on the satellite. A look-back photo faces away from the junction: inbound traffic comes toward the camera, so the driver's left is on the RIGHT of that photo.\par
Rulers give x (top and bottom) and y (right side) in crop units. Dashed magenta lines are stations: S1 y=\emph{y} (10 m from the junction centre), ..., S8 y=\emph{y} (72 m from the junction centre).\par
1. stop\_\allowbreak{}line: where the stop bar crosses the inbound lanes (or, without a painted stop bar, the upstream edge of the crosswalk), as two y values: y\_\allowbreak{}centre where it meets the road-centre side of the inbound lanes and y\_\allowbreak{}curb where it meets their curb side. A stop bar square to the road has the same y at both ends; at a skewed junction they differ. Give it whenever this arm has an inbound carriageway: where paint is hidden, place it where vehicles wait or where the crosswalk begins, and say so in notes. null only if this arm has no inbound carriageway.\par
2. exit\_\allowbreak{}start\_\allowbreak{}y: the y where the outbound lanes begin after the junction (the downstream edge of the crosswalk). null if this arm has no outbound carriageway.\par
3. boundaries: every longitudinal line or edge across the whole road, listed left to right, from the outer curb of the left side to the outer curb of the right side: curbs or pavement edges, lane lines, centre lines, median edges, bicycle-lane lines, buffer or parking edges, and the edges of a track bed or rail-only strip that cars do not use. The rails of a streetcar or light-rail track laid in a traffic lane are not boundaries: that lane, from one lane line to the next, is one strip. For each give its kind and its x on every station; use null at a station where it does not exist (inside the junction, upstream of where a turn pocket opens, or where it ends). Where vehicles, shadow or worn paint hide a line that does exist, interpolate between its visible parts and name those stations in interpolated.\par
4. strips: one entry per pair of neighbouring boundaries (strip k lies between boundary k and k+1), in the same order: which carriageway it belongs to (inbound, outbound, none) and its surface\_\allowbreak{}type where the lanes are read: at the stop line for an inbound strip, just past the exit start for an outbound one. A strip can change along the road: parking or a shoulder upstream can become a turn lane near the stop bar, and turning traffic can enter a bicycle lane to turn right. What it is at the stop bar is what counts. So a strip that is parking upstream and a turn lane at the stop line is a motor\_\allowbreak{}vehicle\_\allowbreak{}lane, and a bicycle lane that turning motor traffic also uses at the stop line is a shared\_\allowbreak{}bicycle\_\allowbreak{}motor\_\allowbreak{}lane; say in its evidence how the strip changes. A right-turn lane often lies between a bicycle lane and the curb near the stop bar; it is part of the carriageway. A lane may be paved differently from its neighbours (brick, concrete or coloured surfacing): tell a lane from a median or buffer by whether it is raised or curbed, whether vehicles stand or drive in it and whether the stop bar and lane lines reach it, not by its colour. A concrete gutter pan (about 0.5-0.8 m wide, often lighter) runs along many curbs: its edge is not a lane line, and the gutter belongs to the curbside lane or shoulder. A lane with rails laid in it, which cars also drive in, is one shared\_\allowbreak{}rail\_\allowbreak{}motor\_\allowbreak{}lane strip. A wide strip with no dividing line is one strip; do not invent lines that are not painted.\par
Do not count opposing traffic as inbound or outbound by its position alone: say which way it runs from arrows, stop bars, vehicles and the road centre.\par
Return JSON only:\par
\texttt{\{"stop\_\allowbreak{}line": \{"y\_\allowbreak{}centre": 140.0,\allowbreak{} "y\_\allowbreak{}curb": 140.0\},\allowbreak{} "exit\_\allowbreak{}start\_\allowbreak{}y": 150.0,\allowbreak{}}\par
\texttt{ "boundaries": [\{"kind": "curb\_\allowbreak{}or\_\allowbreak{}pavement\_\allowbreak{}edge\textbar\allowbreak{}lane\_\allowbreak{}line\textbar\allowbreak{}centre\_\allowbreak{}line\textbar\allowbreak{}median\_\allowbreak{}edge\textbar\allowbreak{}bicycle\_\allowbreak{}line\textbar\allowbreak{}buffer\_\allowbreak{}or\_\allowbreak{}parking\_\allowbreak{}edge\textbar\allowbreak{}rail\_\allowbreak{}edge\textbar\allowbreak{}unknown",\allowbreak{} "xs": [null,\allowbreak{} 212.0,\allowbreak{} 211.5,\allowbreak{} ...],\allowbreak{} "interpolated": ["S6"],\allowbreak{} "visibility": "clear\textbar\allowbreak{}partial\textbar\allowbreak{}occluded"\}],\allowbreak{}}\par
\texttt{ "strips": [\{"carriageway": "inbound\textbar\allowbreak{}outbound\textbar\allowbreak{}none",\allowbreak{} "surface\_\allowbreak{}type": "motor\_\allowbreak{}vehicle\_\allowbreak{}lane\textbar\allowbreak{}shared\_\allowbreak{}rail\_\allowbreak{}motor\_\allowbreak{}lane\textbar\allowbreak{}shared\_\allowbreak{}bicycle\_\allowbreak{}motor\_\allowbreak{}lane\textbar\allowbreak{}bicycle\_\allowbreak{}lane\textbar\allowbreak{}parking\textbar\allowbreak{}median\_\allowbreak{}or\_\allowbreak{}buffer\textbar\allowbreak{}shoulder\textbar\allowbreak{}sidewalk\_\allowbreak{}or\_\allowbreak{}verge\textbar\allowbreak{}rail\_\allowbreak{}or\_\allowbreak{}bus\_\allowbreak{}only\textbar\allowbreak{}unknown",\allowbreak{} "confidence": "high\textbar\allowbreak{}medium\textbar\allowbreak{}low",\allowbreak{} "evidence": "what shows it"\}],\allowbreak{}}\par
\texttt{ "notes": "what limited the trace"\}}
\end{promptbox}

\begin{promptbox}{Blind reading (one request per approach)}
This image shows one approach to a signalised intersection.\par
Left: a satellite crop rotated so that traffic on this approach drives UP toward the intersection; the stop bar and crosswalk are near the top. The road's other carriageway runs down the left side.\par
Right: street views of this approach (\emph{C1: pose, distance from the junction centre, capture date}; ...). A forward photo looks along the direction of travel toward the junction. A look-back photo is taken near the stop bar facing away from the junction: the approach's traffic comes toward the camera, so the driver's left is on the RIGHT of that photo. A street view may be old, taken from the opposite carriageway, or partly blocked; say so when it is.\par
Read the lane layout of THIS approach at the stop bar, counting from the driver's left.\par
\textendash\ Count only lanes open to general motor traffic in the direction of travel. Turn pockets count; a lane shared with a streetcar counts.\par
\textendash\ A strip can change along the road: parking or a shoulder upstream can become a turn lane near the stop bar, and turning traffic can enter a bicycle lane to turn right. What it is at the stop bar is what counts. So a curbside turn lane counts where it opens from parking or a shoulder or lies beside a bicycle lane, and so does a bicycle lane that turning motor traffic also uses at the stop bar.\par
\textendash\ A lane may be paved differently from its neighbours (brick, concrete or coloured surfacing): tell a lane from a median or buffer by whether it is raised or curbed, whether vehicles stand or drive in it and whether the stop bar and lane lines reach it, not by its colour.\par
\textendash\ Do not count opposing lanes, medians, rail- or bus-only lanes, gutters, or bicycle lanes, parking and shoulders that motor traffic does not use at the stop bar; report those separately.\par
\textendash\ For each counted lane give the movements its pavement arrows allow (left, through, right, u\_\allowbreak{}turn), or an empty list if no arrow is legible.\par
\textendash\ If shadow, occlusion, image age or camera position prevents a reliable count, set count\_\allowbreak{}settled to false instead of guessing.\par
Return JSON only:\par
\texttt{\{"lanes":[\{"lane":1,\allowbreak{}"turns":["left"],\allowbreak{}"evidence":"what shows this lane and its arrows"\}],\allowbreak{}"bicycle\_\allowbreak{}lanes":0,\allowbreak{}"rail\_\allowbreak{}or\_\allowbreak{}bus\_\allowbreak{}only\_\allowbreak{}lanes":0,\allowbreak{}"parking\_\allowbreak{}lanes":0,\allowbreak{}"count\_\allowbreak{}confidence":"high\textbar\allowbreak{}medium\textbar\allowbreak{}low",\allowbreak{}"count\_\allowbreak{}settled":true,\allowbreak{}"limitations":"what limited the reading"\}}
\end{promptbox}

The re-trace sends the trace prompt followed by the text below. The separate reading appears only when it counts more lanes than the trace, and its last sentence only when the two differ.

\begin{promptbox}{Re-trace: text added to the trace prompt}
This is a second look. The dashed cyan lines b1..bN on the crop are your earlier boundaries, and the yellow lines your stop line and exit start. The earlier trace gives the inbound carriageway \emph{n} general motor lane(s) at the stop line.\par
A separate reading of this approach from its street views and an unmarked satellite crop, made without seeing your trace, gave a settled count of \emph{n} general motor lane(s) at the stop bar (\emph{arrows}); its own limits: \emph{limitations it stated}. It differs from your trace, so one of the two is wrong.\par
Measured from your trace:\par
\textendash\ \emph{one line per measured issue}\par
In the street views on the right, your boundaries b1..bN are projected onto the road from each camera's recorded position and heading (camera height 2.5 m assumed); the projection can sit up to about a lane to one side or skew with distance. Use it to tell which painted line each boundary stands for: a lane in a photo with no strip of its own between two projected boundaries means a missing boundary; a projected strip over a curb, gutter, sidewalk, bay or the opposing lanes means a spurious one.\par
Look again, using the lightened panel where shadow covers the road: a boundary may be missing, spurious or on the wrong side of the road centre, or a strip may hold the wrong surface. Return the complete corrected answer in the same format. Keep the trace where the pixels support it, even if it still differs from the separate reading or a strip stays unusually wide or narrow.
\end{promptbox}

When lane use reported lanes without an outline, the re-trace adds:

\begin{promptbox}{Re-trace addition: lanes without an outline}
A lane-use check of this approach, which saw your inbound lanes outlined on the satellite and projected into the street views, reported general motor lane(s) at the stop bar that have no strip of their own:\par
\textendash\ \emph{where}: \emph{evidence}\par
Add the boundary that separates such a lane where the pixels show it; keep the trace as it is where they do not.
\end{promptbox}

When no lane allows a turn toward an exit on the left, it adds:

\begin{promptbox}{Re-trace addition: no lane toward a left exit}
The junction has an exit to the left of this approach (\emph{exit link}, heading \emph{compass heading}, turning \emph{signed turn angle} deg), but a lane-use check of this approach, which saw your inbound lanes outlined on the satellite and projected into the street views, found no inbound lane that allows a left turn. Look again at the road-centre side of the inbound carriageway for a left-turn lane or pocket that the trace leaves out, merges into a neighbouring strip or gives another surface type (such as a median or buffer, or a rail- or bus-only strip). Where the pixels show such a lane at the stop line, give it its own strip as a motor lane; where they show none (left turns may share a lane or be banned), keep the trace as it is.
\end{promptbox}

\begin{promptbox}{Lane use (one request per approach)}
Read lane use and turning movements for the \emph{approach} approach of the signalised intersection \emph{intersection}.\par
The image shows: the approach (satellite, traffic drives UP toward the junction, its general motor lanes outlined in yellow and labelled r\#), the same with shadows lightened; a north-up overview of the junction with the lanes of every exit outlined in cyan and labelled exit:r\#; and street views of this approach (\emph{C1: pose, distance from the junction centre, capture date}; ...) with the lanes r\# outlined in yellow where they project onto the road from each camera (camera height assumed; the outline can sit up to about a lane to one side). A look-back photo faces away from the junction: the approach's traffic comes toward the camera, so the driver's left is on the RIGHT of that photo.\par
Approach lanes:\par
\textendash\ r1: lane 1 from the driver's left, about \emph{lane width} m wide\par
\textendash\ ...: lane ... from the driver's left, about \emph{lane width} m wide\par
The trace across the approach at the stop line, from the driver's left: \emph{lanes, bicycle lanes, parking and medians with their widths}.\par
Exits:\par
\textendash\ \emph{exit link}: traffic leaving the junction heading \emph{compass heading}, turning \emph{signed turn angle} deg from this approach (\emph{left, straight ahead, right, sharp or back}); lanes r1, ... from its driver's left (the road centre side)\par
A separate reading of this approach from its street views (made without seeing the lanes r\#) counted \emph{n} general motor lane(s) with arrows \emph{arrows} from the driver's left; its limits: \emph{limitations it stated}. Use it for arrows and signs; where it counts fewer lanes than the r\#, judge each lane from the images, not from that count.\par
1. For every lane r\# first say whether it exists as a general motor lane at the stop bar (yes; no: it is really a gutter, shoulder, parking, bay, a bicycle lane that motor traffic does not use there, sidewalk or part of another lane; uncertain), judged from the satellite and the street views. A strip can change along the road: parking or a shoulder upstream can become a turn lane near the stop bar, and turning traffic can enter a bicycle lane to turn right. What it is at the stop bar is what counts. A lane may be paved differently from its neighbours (brick, concrete or coloured surfacing): tell a lane from a median or buffer by whether it is raised or curbed, whether vehicles stand or drive in it and whether the stop bar and lane lines reach it, not by its colour. Then give the turns it allows (left, through, right, u\_\allowbreak{}turn; none for a lane that does not exist), the basis (observed: an arrow, ONLY word or sign read for this lane; inferred: from lane order, lane lines or the junction layout; unknown) and the evidence. Bind a marking to a lane by its position in the images, not by vehicle positions alone. The curbside lane may also turn right unless a permanent marking or fixed sign rules it out; this applies only to a strip that is a lane. The network records the permanent layout. Cones, barrels, drums, delineator posts, temporary barriers, portable signs on stands or weighted bases and work-zone fencing are temporary traffic control: a lane closed or a turn banned only by them keeps its permanent existence and turns. Judge exists and turns from the permanent markings and fixed signs, and report any temporary control in temporary\_\allowbreak{}control.\par
2. For each turn (left, through, right, u\_\allowbreak{}turn) name the exit it leads to (or none), whether you propose it (candidate), cannot settle it (unresolved) or see no such movement (not\_\allowbreak{}proposed), and its lane pairs: which approach lane feeds which exit lane. A lane that allows a turn must appear in that turn's pairs, and only such lanes may. A pair names lanes by their IDs (approach r\#, exit r\#).\par
3. If a small arrow, word or sign could settle a lane's use, ask for up to three native-resolution crops in detail\_\allowbreak{}requests (source: approach, overview, or a street-view ID; bbox\_\allowbreak{}xyxy as 0-1 fractions of that panel). Otherwise leave detail\_\allowbreak{}requests empty.\par
4. missing\_\allowbreak{}lanes: a general motor lane you see at the stop bar that has no r\# of its own (for example a turn pocket or a curbside turn lane the outlines leave out, or two lanes inside one outline): where it lies (for example ``between r4 and the curb'') and what shows it. Leave it empty when every lane has its own r\#.\par
5. temporary\_\allowbreak{}control: whether you see temporary traffic control on this approach (yes, no, uncertain), which r\# it affects, what it does to them (for example closes a lane or bans a turn for now), and what shows it in which image. The lanes and turns above stay the permanent ones.\par
Return JSON only:\par
\texttt{\{"lanes": [\{"region\_\allowbreak{}id": "r1",\allowbreak{} "exists": "yes\textbar\allowbreak{}no\textbar\allowbreak{}uncertain",\allowbreak{} "turns": ["left"],\allowbreak{} "basis": "observed\textbar\allowbreak{}inferred\textbar\allowbreak{}unknown",\allowbreak{} "confidence": "high\textbar\allowbreak{}medium\textbar\allowbreak{}low",\allowbreak{} "evidence": "..."\}],\allowbreak{}}\par
\texttt{ "movements": [\{"turn": "left",\allowbreak{} "to": "out\_\allowbreak{}NB",\allowbreak{} "status": "candidate\textbar\allowbreak{}unresolved\textbar\allowbreak{}not\_\allowbreak{}proposed",\allowbreak{} "lane\_\allowbreak{}pairs": [["r1",\allowbreak{} "r1"]],\allowbreak{} "reason": "..."\}],\allowbreak{}}\par
\texttt{ "detail\_\allowbreak{}requests": [\{"source": "approach",\allowbreak{} "bbox\_\allowbreak{}xyxy": [0.1,\allowbreak{} 0.0,\allowbreak{} 0.4,\allowbreak{} 0.3],\allowbreak{} "why": "..."\}],\allowbreak{}}\par
\texttt{ "missing\_\allowbreak{}lanes": [\{"where": "between r4 and the curb",\allowbreak{} "evidence": "..."\}],\allowbreak{}}\par
\texttt{ "temporary\_\allowbreak{}control": \{"seen": "yes\textbar\allowbreak{}no\textbar\allowbreak{}uncertain",\allowbreak{} "lanes": ["r2"],\allowbreak{} "effect": "...",\allowbreak{} "evidence": "..."\},\allowbreak{}}\par
\texttt{ "notes": "..."\}}
\end{promptbox}

\paragraph{Notes of the lane-use prompt.} Each is added after the separate reading only where code finds its condition. A lane at least 2.5\,m wide between a bicycle lane and the curb:

\begin{promptbox}{Lane-use note: wide curbside lane}
\emph{r\#} (\emph{width} m) lies between a bicycle lane and the curb and is as wide as a motor lane (2.5 m or more at the stop line). Curbside parking ends short of a junction, so such a strip is normally where right-turning traffic waits at the stop bar, even where cars park in it upstream and no arrow is painted. Judge it not a general motor lane only if something at the stop bar keeps cars out of it (a curb extension or island, posts, a bus stop, hatching, or a stop bar that ends short of it), and say what.
\end{promptbox}

A bicycle strip at least 2.5\,m wide between two lanes:

\begin{promptbox}{Lane-use note: wide bicycle strip}
The trace has a bicycle lane \emph{width} m wide between \emph{r\#} and \emph{r\#}, wider than a bicycle lane usually is: if a general motor lane (for example a turn lane) lies within that strip at the stop bar, report it in missing\_\allowbreak{}lanes; otherwise leave it.
\end{promptbox}

A lane given left and through by inference while another lane allows through:

\begin{promptbox}{Lane-use note: shared lane by inference}
An earlier answer gave \emph{r\#} both left and through, inferred rather than read from an arrow or sign, while other lanes also carry through traffic. Where an approach has more than one through lane, the lane on the road-centre side that takes left turns is usually a left-turn lane, and a left-turn lane may lie on the road-centre side outside the outlines, on paving that looks like a median. Look again: give it only the turns that arrows, signs or position show, and report in missing\_\allowbreak{}lanes a left-turn lane that has no outline of its own.
\end{promptbox}

A left exit that no lane serves after a re-trace found no left-turn lane:

\begin{promptbox}{Lane-use note: left exit with no lane}
An exit lies to the left of this approach (\emph{exit link}, heading \emph{compass heading}), but an earlier answer gave no lane a left turn, and a second trace found no further lane on the road-centre side. Look again at the leftmost lane: say whether it also allows left turns (for example a lane shared by left and through traffic) or whether a sign or marking bans left turns here, and give left where nothing rules it out.
\end{promptbox}

Lanes an earlier answer removed while the blind reading does not count fewer lanes:

\begin{promptbox}{Lane-use note: removed lanes kept}
An earlier answer judged \emph{r\#} not to be a general motor lane at the stop bar. \emph{one sentence on what the separate reading gave} So nothing independent of the outlines confirms that, and it stays a lane: treat it as a lane (exists yes, or uncertain where the images cannot settle it) and give the turns it allows.
\end{promptbox}

A lane at least 2.5\,m wide that an earlier answer removed:

\begin{promptbox}{Lane-use note: wide lane removed}
An earlier answer judged \emph{r\#} (\emph{width} m) not to be a general motor lane, but it is as wide as a motor lane at the stop line (2.5 m or more). Keep it as a lane unless something at the stop bar keeps cars out (a curb extension or island, posts, a bus stop, hatching, or a stop bar that ends short of it); if so, say what.
\end{promptbox}

\begin{promptbox}{Detail crops: text added to the lane-use prompt}
This is a second look. The image now shows the native-resolution crops you asked for, side by side, each titled with its source and box. Your earlier answer was:\par
\emph{earlier answer}\par
Return the complete answer again in the same format, revised where the crops change it, with detail\_\allowbreak{}requests empty.
\end{promptbox}

\paragraph{Single-request prompt.} The satellite variant sends, with the north-up satellite image:
\begin{promptbox}{Single request (satellite variant)}
This is a satellite image of the signalised intersection \emph{name}. North is up. Approaches to the intersection:\par
\textendash\ EB: traffic heading east arriving from the west arm [one line per approach].\par
For each approach, count the motor-vehicle lanes at the stop line and give each lane's use, ordered from the driver's left. Lane uses: L, T, R, U or combinations such as LT, TR, LTR. Return JSON only:\par
\{"approaches": [\{"direction": "EB", "lanes": ["L", "T", "TR"], "confidence": "low\textbar medium\textbar high", "note": "..."\}]\}
\end{promptbox}
The variant with street views replaces the first two sentences with: ``This is one image of the signalised intersection \emph{name}. Left: a satellite image, north up, centred on the junction. Right: street-level photos taken on the approaches, one row per approach (EB, NB, \dots, top to bottom). Each photo is labelled with its approach, position, distance from the junction, capture date and the way the camera looks.''

\paragraph{Editing prompts.} Every editing request sends the system prompt below, a fixed instruction that precedes the request, with one image and one query. The system prompt is the same for all scenarios, and every query has the same parts: the brief, what the image shows, the network facts, closed catalogues, the task and the answer format. The content of each part depends on the scenario type, since a lane addition, an interchange conversion and a passing lane differ in the network shown, the parts to bind and the answer (Table~\ref{tab:scenarios}). The interchange query is the editing package's own: code fills it from the brief and the network tables, and its catalogues list every form the templates build. The lane-addition and passing-lane queries were written for their types and have each served one scenario. The queries are printed as the runs sent them, the same in every run and for both models; italics replace the rows of the tables.

\begin{promptbox}{System prompt (editing, all scenarios)}
You are the proposer stage of a traffic-network scenario-editing pipeline.\par
You never draw, digitize or estimate coordinates. You answer only with IDs that exist in the tables you are given and with labels from the closed catalogs. Topology, turning movements and geometry are produced afterwards by deterministic templates and checked by code; anything you name that is not in the tables is rejected. Return exactly one JSON object and nothing else.
\end{promptbox}

\begin{promptbox}{S1 query: lane addition}
\promptsec{SCENARIO}\par
L303 \& US-60 interchange improvements: add one general-purpose lane in each direction (from two lanes to three) on US-60 (Grand Avenue) for about half a mile on its approach to the Loop 303 interchange from the northwest, ending where the interchange begins, in the AM, MD and PM periods.\par
\promptsec{IMAGE}\par
Panel A (left): the whole meso subarea, north up; the red box is the area of panel B.\par
Panel B (right): the meso network inside the red box, north up, in a local frame in metres (x east, y north, origin at the centre of the panel). Each two-way road is drawn once as a line between its two nodes and labelled with its number; the link table lists its two directed links \textless{}number\textgreater{}AB and \textless{}number\textgreater{}BA. Colours: grey freeway, blue arterial, light blue collector, green ramp; dashed lines are centroid connectors (unlabelled). A 6x6 grid is laid over panel B: rows A-F top to bottom, columns 1-6 left to right.\par
\promptsec{NODE TABLE} (node\_\allowbreak{}id,x,y) for the nodes in panel B\par
\emph{36 rows, one per node}\par
\promptsec{LINK TABLE} (link\_\allowbreak{}id,from\_\allowbreak{}node\_\allowbreak{}id,to\_\allowbreak{}node\_\allowbreak{}id,link\_\allowbreak{}type,lanes,corridor,length\_\allowbreak{}mi); all links are directed; corridor is the traffic-message-channel corridor name where the link has one\par
\emph{58 rows, one per link}\par
\promptsec{CLOSED CATALOGS}\par
operations: add\_\allowbreak{}lane, remove\_\allowbreak{}lane, add\_\allowbreak{}link, change\_\allowbreak{}speed, add\_\allowbreak{}ramp\par
periods: AM, MD, PM\par
expected\_\allowbreak{}facts vocabulary:\par
\textendash\ both\_\allowbreak{}directions\_\allowbreak{}widened: both directions of the named road get the lane\par
\textendash\ topology\_\allowbreak{}unchanged: no node or link is added or removed\par
\textendash\ only\_\allowbreak{}named\_\allowbreak{}links\_\allowbreak{}change: no link outside the answer changes\par
\textendash\ capacity\_\allowbreak{}scales\_\allowbreak{}with\_\allowbreak{}lanes: link capacity grows with the lane count\par
\textendash\ all\_\allowbreak{}zone\_\allowbreak{}pairs\_\allowbreak{}preserved: every zone pair connected before stays connected\par
\promptsec{TASK}\par
1. Find the road the request names, and the stretch of it the request describes, in panel B and the tables.\par
2. List every directed link to widen, in both directions. For every link copy its from\_\allowbreak{}node\_\allowbreak{}id and to\_\allowbreak{}node\_\allowbreak{}id and its current lanes from the LINK TABLE, and give the lanes after the edit.\par
3. Give the grid cell of panel B that contains the middle of the stretch.\par
4. List the expected\_\allowbreak{}facts (from the vocabulary) that must hold after the edit.\par
5. List what the image and tables do not establish in not\_\allowbreak{}established.\par
\promptsec{OUTPUT}: one JSON object with the keys in exactly this order\par
\texttt{\{}\par
\texttt{  "evidence": ["\textless{}short observations from the image and tables,\allowbreak{} written before any decision\textgreater{}"],\allowbreak{}}\par
\texttt{  "facility": "\textless{}the road the request names,\allowbreak{} as in the corridor column without its direction\textgreater{}",\allowbreak{}}\par
\texttt{  "anchor\_\allowbreak{}cell": "\textless{}grid cell\textgreater{}",\allowbreak{}}\par
\texttt{  "links": [\{"link\_\allowbreak{}id": "\textless{}id\textgreater{}",\allowbreak{} "from\_\allowbreak{}node\_\allowbreak{}id": "\textless{}id\textgreater{}",\allowbreak{} "to\_\allowbreak{}node\_\allowbreak{}id": "\textless{}id\textgreater{}",\allowbreak{} "lanes\_\allowbreak{}before": \textless{}int\textgreater{},\allowbreak{} "lanes\_\allowbreak{}after": \textless{}int\textgreater{}\}],\allowbreak{}}\par
\texttt{  "operations": [\{"op": "\textless{}operation\textgreater{}",\allowbreak{} "lanes\_\allowbreak{}added": \textless{}int\textgreater{}\}],\allowbreak{}}\par
\texttt{  "periods": ["\textless{}period\textgreater{}",\allowbreak{} ...],\allowbreak{}}\par
\texttt{  "expected\_\allowbreak{}facts": ["\textless{}fact\textgreater{}",\allowbreak{} ...],\allowbreak{}}\par
\texttt{  "not\_\allowbreak{}established": ["\textless{}free text\textgreater{}",\allowbreak{} ...]}\par
\texttt{\}}
\end{promptbox}

\begin{promptbox}{S2 query: interchange conversion}
\promptsec{SCENARIO}\par
SR 347 / I-10 (Queen Creek Rd interchange): convert the conventional diamond interchange to a diverging diamond interchange (DDI).\par
\promptsec{IMAGE}\par
The image has two panels.\par
Panel A (left): satellite image of the site BEFORE the project (Mapbox imagery, no network drawn on it).\par
Panel B (right): the GMNS base network rendered in a schematic local coordinate system.\par
Coordinates in metres on a local tangent plane centred on the interchange: x increases to the east, y to the north (north up). Panel A is the same area, north up, at about the same scale.\par
Link IDs are printed in coloured boxes (grey freeway, blue arterial, green ramp); intersection connectors are orange and unlabelled. Node IDs are printed next to the white circles.\par
A 6x6 grid is overlaid on panel B: rows A-F top to bottom, columns 1-6 left to right (cell ``C2'' = third row, second column).\par
\promptsec{NODE TABLE} (node\_\allowbreak{}id,name,node\_\allowbreak{}type,x,y)\par
\emph{42 rows, one per node}\par
\promptsec{LINK TABLE} (link\_\allowbreak{}id,name,from\_\allowbreak{}node\_\allowbreak{}id,to\_\allowbreak{}node\_\allowbreak{}id,facility\_\allowbreak{}type); all links are directed\par
\emph{47 rows, one per link}\par
\promptsec{CLOSED CATALOGS}\par
facility\_\allowbreak{}class: diamond, tight\_\allowbreak{}diamond, ddi, spui, roundabout\_\allowbreak{}diamond, parclo\_\allowbreak{}a, parclo\_\allowbreak{}b, parclo\_\allowbreak{}ab, cloverleaf, directional, trumpet, other\_\allowbreak{}or\_\allowbreak{}unclear\par
operations: replace\_\allowbreak{}facility, add\_\allowbreak{}lane, add\_\allowbreak{}ramp, add\_\allowbreak{}flyover, realign\_\allowbreak{}ramp, add\_\allowbreak{}managed\_\allowbreak{}lanes, meter\_\allowbreak{}ramps, set\_\allowbreak{}allowed\_\allowbreak{}uses\par
terminal roles (one host link per role, per ramp terminal):\par
\textendash\ approach\_\allowbreak{}in: arterial link arriving at this terminal from its outer leg (the side away from the freeway) (facility\_\allowbreak{}type must be arterial)\par
\textendash\ departure\_\allowbreak{}out: arterial link leaving this terminal toward its outer leg (facility\_\allowbreak{}type must be arterial)\par
\textendash\ bridge\_\allowbreak{}in: arterial link arriving at this terminal from the other terminal, across the freeway (facility\_\allowbreak{}type must be arterial)\par
\textendash\ bridge\_\allowbreak{}out: arterial link leaving this terminal toward the other terminal, across the freeway (facility\_\allowbreak{}type must be arterial)\par
\textendash\ offramp\_\allowbreak{}in: off-ramp link that ends at this terminal (facility\_\allowbreak{}type must be ramp)\par
\textendash\ onramp\_\allowbreak{}out: on-ramp link that starts at this terminal (facility\_\allowbreak{}type must be ramp)\par
expected\_\allowbreak{}facts vocabulary:\par
\textendash\ all\_\allowbreak{}external\_\allowbreak{}od\_\allowbreak{}preserved: every external origin-destination pair reachable before the edit stays reachable\par
\textendash\ freeway\_\allowbreak{}links\_\allowbreak{}unchanged: no freeway mainline link changes\par
\textendash\ six\_\allowbreak{}terminal\_\allowbreak{}movements\_\allowbreak{}per\_\allowbreak{}terminal: each terminal keeps its 6 movements and gains no U-turn or ramp-to-ramp movement\par
\textendash\ through\_\allowbreak{}movements\_\allowbreak{}cross\_\allowbreak{}over\_\allowbreak{}at\_\allowbreak{}each\_\allowbreak{}terminal: both arterial through movements pass a crossover node at each terminal\par
\textendash\ left\_\allowbreak{}side\_\allowbreak{}running\_\allowbreak{}between\_\allowbreak{}terminals: between the terminals, arterial traffic runs on the left side\par
\textendash\ no\_\allowbreak{}collateral\_\allowbreak{}changes: nothing outside the bound terminals and the links between them changes\par
\textendash\ ten\_\allowbreak{}spui\_\allowbreak{}movements: (SPUI) the single point carries the 10 through, left and right movements and no U-turn or ramp-to-ramp movement\par
\textendash\ single\_\allowbreak{}signal\_\allowbreak{}controls\_\allowbreak{}all\_\allowbreak{}ramp\_\allowbreak{}movements: (SPUI) one signal controls every ramp and cross-street movement\par
\textendash\ all\_\allowbreak{}entries\_\allowbreak{}reach\_\allowbreak{}all\_\allowbreak{}exits: (roundabout terminals) every entry reaches every exit, U-turns included\par
\textendash\ counterclockwise\_\allowbreak{}circulation: (roundabout terminals) each ring circulates counter-clockwise\par
\textendash\ yield\_\allowbreak{}controlled\_\allowbreak{}entries: (roundabout terminals) entries yield; no signals on the rings\par
\promptsec{TASK}\par
1. Read the facility class of the existing interchange from panel A and the tables.\par
2. Identify each ramp terminal (the at-grade junction where ramps meet the arterial). Name it by its compass side along the arterial (for example ``SW'', ``NE'') and give the grid cell that contains it.\par
3. For each terminal bind every role to exactly one link\_\allowbreak{}id from the LINK TABLE. For every link you name, copy its from\_\allowbreak{}node\_\allowbreak{}id and to\_\allowbreak{}node\_\allowbreak{}id from the table; a mismatch rejects the proposal.\par
4. Write the operation list that the scenario sentence asks for, using only catalog operations.\par
5. List the expected\_\allowbreak{}facts (from the vocabulary) that must hold after the edit.\par
6. List everything the image and text do not establish (lane counts, signal timing, turn pockets, demand, ...) in not\_\allowbreak{}established. Unknown is not zero.\par
\promptsec{OUTPUT}: one JSON object with the keys in exactly this order\par
\texttt{\{}\par
\texttt{  "evidence": ["\textless{}short observations from the image and tables,\allowbreak{} written before any decision\textgreater{}"],\allowbreak{}}\par
\texttt{  "facility\_\allowbreak{}class": "\textless{}facility\_\allowbreak{}class from the catalog\textgreater{}",\allowbreak{}}\par
\texttt{  "arterial\_\allowbreak{}axis\_\allowbreak{}nodes": [\textless{}boundary node id at one end of the arterial\textgreater{},\allowbreak{} \textless{}boundary node id at the other end\textgreater{}],\allowbreak{}}\par
\texttt{  "terminals": [}\par
\texttt{    \{"name": "\textless{}side\textgreater{}",\allowbreak{} "anchor\_\allowbreak{}cell": "\textless{}grid cell\textgreater{}",\allowbreak{}}\par
\texttt{      "bindings": \{"\textless{}role\textgreater{}": \{"link\_\allowbreak{}id": \textless{}int\textgreater{},\allowbreak{} "from\_\allowbreak{}node\_\allowbreak{}id": \textless{}int\textgreater{},\allowbreak{} "to\_\allowbreak{}node\_\allowbreak{}id": \textless{}int\textgreater{}\},\allowbreak{} ...all 6 roles...\}\}}\par
\texttt{  ],\allowbreak{}}\par
\texttt{  "operations": [\{"op": "\textless{}operation\textgreater{}",\allowbreak{} "from\_\allowbreak{}class": "\textless{}facility\_\allowbreak{}class\textgreater{}",\allowbreak{} "to\_\allowbreak{}class": "\textless{}facility\_\allowbreak{}class\textgreater{}",\allowbreak{} "terminals": ["\textless{}side\textgreater{}",\allowbreak{} ...]\}],\allowbreak{}}\par
\texttt{  "expected\_\allowbreak{}facts": ["\textless{}fact\textgreater{}",\allowbreak{} ...],\allowbreak{}}\par
\texttt{  "not\_\allowbreak{}established": ["\textless{}free text\textgreater{}",\allowbreak{} ...]}\par
\texttt{\}}
\end{promptbox}

\begin{promptbox}{S3 query: passing lanes}
\promptsec{SCENARIO}\par
US-60 passing lanes (2022 ADOT DTA briefing, scenario 3): add a passing lane in the EB direction from MP 349 to MP 350 and in the WB direction from MP 350 to MP 351, one mile each. Mileposts on US-60 increase eastward. Lane use (briefing slide 4): passing lane - regular autos yes, slow vehicles no; general lane - regular autos yes, slow vehicles yes.\par
\promptsec{IMAGE}\par
Panel A (top): the location figure of the 2022 briefing deck for this scenario, slide 3 ``Current Configuration'': a satellite picture of US-60 with the deck's two call-outs. It is a picture of the place, not of the network.\par
Panel B (bottom): the No-Build micro network between the two traffic zones at the ends of the study corridor, north up.\par
The blue line is the eastbound (EB) carriageway of US-60, the green line the westbound (WB) carriageway (they run side by side); grey lines are side roads and the roads beyond the zones. Stations are placed every 100 m along the corridor, measured from the west zone: station E\#\#\# lies on the EB carriageway and W\#\#\# on the WB carriageway, \#\#\# = tenths of a kilometre (E047 = EB carriageway 4.7 km from the west zone). Every 10th station is labelled; ticks mark every 100 m.\par
Red dots are junctions where a side road joins US-60.\par
\promptsec{CORRIDOR FACTS} (from the network)\par
length between the zones: 14.55 km; stations E000-E145 and W000-W145\par
lanes: one lane in each direction on the whole corridor\par
\promptsec{JUNCTIONS} (id,km,station)\par
J1,6.79,E067/W067\par
\promptsec{CLOSED CATALOGS}\par
directions: EB, WB\par
vehicle classes: auto, slow\_\allowbreak{}vehicle, truck (slow vehicles drive at 85 \% of the auto speed, trucks at 50 \%)\par
expected\_\allowbreak{}facts vocabulary:\par
\textendash\ one\_\allowbreak{}lane\_\allowbreak{}today: each section has one lane in its direction before the edit\par
\textendash\ no\_\allowbreak{}junction\_\allowbreak{}inside: no side road joins inside a section\par
\textendash\ passing\_\allowbreak{}lane\_\allowbreak{}auto\_\allowbreak{}only: only autos may use the passing lane\par
\textendash\ general\_\allowbreak{}lane\_\allowbreak{}all\_\allowbreak{}classes: every vehicle class may use the general lane\par
\textendash\ existing\_\allowbreak{}cells\_\allowbreak{}unchanged: existing nodes and links keep their IDs and positions\par
\textendash\ zones\_\allowbreak{}connected\_\allowbreak{}for\_\allowbreak{}all\_\allowbreak{}classes: zone 1 and zone 2 stay connected both ways for every class\par
\promptsec{TASK}\par
1. From panel A, read where the deck places the two passing lanes, and find that place in panel B.\par
2. For each direction give the station where its passing lane begins and the station where it ends (west and east), consistent with the mileposts in the request.\par
3. Give the vehicle classes allowed on the passing lane and on the general lane.\par
4. List the expected\_\allowbreak{}facts (from the vocabulary) that must hold after the edit.\par
5. List what the image and text do not establish in not\_\allowbreak{}established.\par
\promptsec{OUTPUT}: one JSON object with the keys in exactly this order\par
\texttt{\{}\par
\texttt{  "evidence": ["\textless{}short observations from the panels and facts,\allowbreak{} written before any decision\textgreater{}"],\allowbreak{}}\par
\texttt{  "sections": [}\par
\texttt{    \{"direction": "EB",\allowbreak{} "mileposts": "349-350",\allowbreak{} "west\_\allowbreak{}station": "E\#\#\#",\allowbreak{} "east\_\allowbreak{}station": "E\#\#\#"\},\allowbreak{}}\par
\texttt{    \{"direction": "WB",\allowbreak{} "mileposts": "350-351",\allowbreak{} "west\_\allowbreak{}station": "W\#\#\#",\allowbreak{} "east\_\allowbreak{}station": "W\#\#\#"\}}\par
\texttt{  ],\allowbreak{}}\par
\texttt{  "lane\_\allowbreak{}use": \{"passing\_\allowbreak{}lane": ["\textless{}class\textgreater{}",\allowbreak{} ...],\allowbreak{} "general\_\allowbreak{}lane": ["\textless{}class\textgreater{}",\allowbreak{} ...]\},\allowbreak{}}\par
\texttt{  "expected\_\allowbreak{}facts": ["\textless{}fact\textgreater{}",\allowbreak{} ...],\allowbreak{}}\par
\texttt{  "not\_\allowbreak{}established": ["\textless{}free text\textgreater{}",\allowbreak{} ...]}\par
\texttt{\}}
\end{promptbox}

\paragraph{Repair.} Sent only when a gate rejects the first answer, and at most once. In our runs only GPT's three first S3 answers were rejected. Each began the westbound section at W067 (km 6.70), the station the junction table lists for J1, which lies at km 6.79; the side road thus joins 90\,m inside the section, beyond the 50\,m that P6 allows. Each repair moved the section to W068--W084, where Claude's answers had put it. The request, with its one defect:

\begin{promptbox}{Repair request}
\emph{the query}\par
\promptsec{YOUR PREVIOUS ANSWER}\par
\emph{previous answer}\par
\promptsec{THE CHECKS REJECTED IT FOR THESE REASONS}\par
\textendash\ [P6 no junction] WB: junction J1 (km 6.79) lies inside the section\par
Return a corrected JSON object with the same keys in the same order. Fix only what the defects name; re-read the tables for every ID you use.
\end{promptbox}

\paragraph{General agent (no harness).} Each run gets this page; the brief is the scenario and image parts of PAVE's query:

\begin{promptbox}{General agent: the task page}
TASK = \emph{task folder}\par
You are a transport modeller. Edit the road network in TASK/network/ as TASK/brief.md asks, and write the complete Build network.\par
\promptsec{What is in TASK}\par
\textendash\ brief.md: the planning request, and what the image shows.\par
\textendash\ image.png: the image that comes with the request.\par
\textendash\ network/: the No-Build network files \emph{listed for the scenario}.\par
\textendash\ conventions.md: how the network files encode the road.\par
\promptsec{What to deliver, in TASK/output/}\par
\textendash\ The complete Build network as \emph{the same files}, with the same columns and formats as the input files.\par
\textendash\ notes.md: what you changed and how, and anything you could not establish or are unsure of.\par
\promptsec{Requirements}\par
\textendash\ Keep every existing ID. Keep existing nodes and links (their ends, geometry and attributes) as they are unless the edit requires changing them, and give new nodes and links new IDs.\par
\textendash\ Change nothing outside the project the brief describes.\par
\textendash\ Every trip that is possible in the No-Build network must still be possible.\par
\promptsec{How to work}\par
\textendash\ Start with \texttt{cd TASK}. You may read every file in TASK, look at the image, write and run Python scripts in TASK (\emph{Python interpreter}; pandas, numpy, scipy and shapely are installed) and write files in TASK.\par
\textendash\ Open, list or search nothing outside TASK, use no web access and install nothing.\par
\textendash\ Nobody reviews your result and there is no second round: what is in TASK/output/ when you stop is the result.\par
\textendash\ When you are done, reply with one line: ``done \emph{task folder name}'', and anything that went wrong.
\end{promptbox}

\end{document}